\documentclass{article}

\usepackage{microtype}
\usepackage{graphicx}
\usepackage{subfigure}
\usepackage{booktabs} 

\usepackage{hyperref}

\usepackage[accepted]{icml2024}

\usepackage{amsmath}
\usepackage{amssymb}
\usepackage{mathtools}
\usepackage{amsthm}
\usepackage{bbm}
\usepackage{bm}
\usepackage{makecell}
\usepackage{wrapfig}
\usepackage{booktabs}
\usepackage{multirow}
\usepackage{graphicx}
\usepackage{enumitem}
\usepackage{cuted}
\usepackage{capt-of}
\usepackage{caption}
\usepackage{tablefootnote}
\usepackage{threeparttable}

\usepackage[capitalize,noabbrev]{cleveref}

\theoremstyle{plain}

\theoremstyle{definition}

\theoremstyle{remark}

\renewcommand\algorithmiccomment[1]{\hfill\% #1}
\renewcommand{\algorithmicrequire}{\textbf{Input:}}
\renewcommand{\algorithmicensure}{\textbf{Output:}}

\usepackage[textsize=tiny]{todonotes}

\icmltitlerunning{\textsc{OxyGenerator}: Reconstructing Global Ocean Deoxygenation Over a Century with Deep Learning}

\begin{document}

\twocolumn[
\icmltitle{\textsc{OxyGenerator}: Reconstructing Global Ocean Deoxygenation \\Over a Century with Deep Learning}




\icmlsetsymbol{equal}{*}

\begin{icmlauthorlist}
\icmlauthor{Bin Lu}{sjtu-ee}
\icmlauthor{Ze Zhao}{sjtu-ee}
\icmlauthor{Luyu Han}{sjtu-ocean}
\icmlauthor{Xiaoying Gan}{sjtu-ee}
\icmlauthor{Yuntao Zhou}{sjtu-ocean}
\icmlauthor{Lei Zhou}{sjtu-ocean}
\icmlauthor{Luoyi Fu}{sjtu-cs}
\icmlauthor{Xinbing Wang}{sjtu-ee,sjtu-cs}
\icmlauthor{Chenghu Zhou}{cas}
\icmlauthor{Jing Zhang}{sjtu-ocean}
\end{icmlauthorlist}

\icmlaffiliation{sjtu-ee}{Department of Electronic Engineering, Shanghai Jiao Tong University, Shanghai, China}
\icmlaffiliation{sjtu-ocean}{School of Oceanography, Shanghai Jiao Tong University, Shanghai, China}
\icmlaffiliation{sjtu-cs}{Department of Computer Science, Shanghai Jiao Tong University, Shanghai, China}
\icmlaffiliation{cas}{Institute of Geographic Sciences and Natural Resources Research, Chinese Academy of Sciences, Beijing, China}

\icmlcorrespondingauthor{Xiaoying Gan}{ganxiaoying@sjtu.edu.cn}

\icmlkeywords{Machine Learning, ICML}

\vskip 0.3in
]



\printAffiliationsAndNotice{}  

\begin{abstract}
Accurately reconstructing the global ocean deoxygenation over a century is crucial for assessing and protecting marine ecosystem. Existing expert-dominated numerical simulations fail to catch up with the dynamic variation caused by global warming and human activities. Besides, due to the high-cost data collection, the historical observations are severely sparse, leading to big challenge for precise reconstruction. In this work, we propose \textsc{OxyGenerator}, the first deep learning based model, to reconstruct the global ocean deoxygenation from 1920 to 2023. Specifically, to address the heterogeneity across large temporal and spatial scales, we propose zoning-varying graph message-passing to capture the complex oceanographic correlations between missing values and sparse observations. Additionally, to further calibrate the uncertainty, we incorporate inductive bias from dissolved oxygen (DO) variations and chemical effects. Compared with in-situ DO observations, \textsc{OxyGenerator} significantly outperforms CMIP6 numerical simulations, reducing MAPE by 38.77\%, demonstrating a promising potential to understand the “breathless ocean” in data-driven manner.
\end{abstract}


\begin{figure*}
    \centering
    \includegraphics[width=\linewidth]{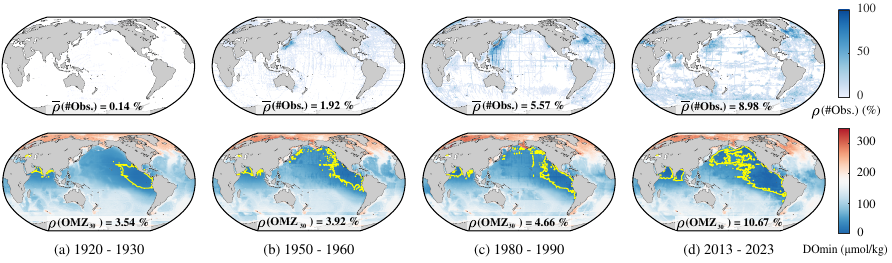}
\captionof{figure}{Global ocean deoxygenation reconstruction via \textsc{OxyGenerator} from 1920 to 2023 based on sparse observation. \textbf{Above}: The proportion of ocean data observed in four-dimensional coordinates $\rho(\text{\#.Obs})$. 
Overall, dissolved oxygen (DO) observation are very sparse in each interval, and many areas have no observations. \textbf{Below}: Minimum DO reconstructed by \textsc{OxyGenerator}. The yellow line envelopes the oxygen minimum zone (OMZ) where $\text{DO}_{\text{min}} \leq 30 \mu \text{mol/kg}$. $\rho(\text{OMZ}_{30})$ indicates the proportion of OMZ30 regions to global oceans, which clearly shows a significant increase over a century.
    \label{figure1}}
\end{figure*}

\section{Introduction}
Oxygen is fundamentally essential for all life. 
Unfortunately, recent research~\cite{Schmidtko2017DeclineIG} has shown that the concentration of dissolved oxygen (DO) in the ocean has been steadily decreasing over the past 50 years, indicating the acceleration of global ocean deoxygenation.
“Breathless ocean” has led to large-scale death of fish, seriously affecting the marine ecosystem~\cite{Cheung2013ShrinkingOF,doi:10.1126/science.aam7240}. Moreover, global warming and human activities further intensify the expansion of oxygen minimum zones in the ocean. 
What is the evolution of global ocean deoxygenation?
Oceanographers eagerly seek a precise understanding of its changes and underlying patterns.

In recent years, due to the advancement of ocean observation technologies and international ocean discovery programs, a lot of in-situ observation data has been accumulated. Some data-driven studies have analyzed the occurrence of ocean deoxygenation in regions such as Atlantic Canada~\cite{hussain2021data}, Southern Ocean~\cite{Giglio2018EstimatingOI}, and Ganga River~\cite{10.1007/978-3-031-24378-3_7}. However, these studies are confined to a local region with short-term observations, typically spanning only several decades. 
The lack of research on long-term global ocean deoxygenation lies in the \emph{severely sparse historical observations} of DO due to the high-cost and high-risk marine scientific expeditions. 
For example, the World Ocean Database (WOD)~\cite{wod2018} is world's largest and widely-used collection of publicly available ocean profile data\footnote[2]{We follow the same method of data gridding in \cite{WOA18DO} with $1^\circ \times 1^\circ$ spatial resolution, 1-year temporal resolution and 0-5500 meters (33 depth levels) of the global ocean.}, and we are extremely surprised to find that more than 96.265\% dissolved oxygen observation data are missing in the past 100 years, largely higher than the missing rate of existing data imputation studies~\cite{DBLP:journals/tkde/MiaoWCGY23,DBLP:journals/csur/AdhikariJZHRAK23}.

\textbf{Prior Works.} To quantitatively understand and predict the long-term trend of global ocean deoxygenation from sparse observations, existing works are mainly categorized into two ways: 
(1) \emph{Numerical Simulation Models}. These methods simulate the DO concentration based on climate models without utilizing in-situ DO observations. Coupled Model Intercomparison Project Phase 6 (CMIP6)~\cite{eyring2016overview}, a world-wide simulation compasrion project, includes different expert-dominated DO simulation over a century.
Many oceanography studies~\cite{Long2021SimulationsWT,Koelling2023DecadalVO,Huang2023ReconstructionOD} often directly leverage these simulation data for long-term analysis or comparisons. However, these models are unable to adjust for DO simulation biases caused by global warming and human activities, leading to error propagation and showing large discrepancies with in-situ observations.
(2) \emph{Spatial Interpolation Methods}. Apart from spatio-temporal numerical simulations, oceanographers utilize spatial regression to interpolate unobserved points based on distance for global analysis. 
\citeauthor{Schmidtko2017DeclineIG} (\citeyear{Schmidtko2017DeclineIG}) adopt distance based weighted average and first quantitatively calculate that oxygen in the ocean has decreased by about 2\% since 1960. 
\citeauthor{Zhou2022ResponsesOH} (\citeyear{Zhou2022ResponsesOH}) further analyze the expanding oceanic oxygen minimum zones with geostatistical regression.
However, these methods only conduct the smoothing of observations through spatial distance, without considering the impact of temporal information and associated factors. 
Therefore, this suggests the following

\textbf{Question:} \emph{Can deep learning methods more accurately reconstruct global ocean deoxygenation over a century under sparse dissolved oxygen observations?}

\textbf{Challenges.} This research question drives the design of specific deep learning methods for reconstructing global ocean deoxygenation. Fortunately, despite the severe sparse observations, the missing values contains implicit spatio-temporal correlations with neighboring observations. Meanwhile, multiple physical-biogeochemical factors show strong connections with dissolved oxygen. However, as a double-edged sword, accurately characterizing the complex correlations between missing values and sparse observations involves two main challenges: (1) \emph{Irregular 4D spatio-temporal heterogeneity}. Dynamic ocean is a four-dimensional irregular spatio-temporal area that includes longitude, latitude, depth, and time. Constrained by both tectonic plates and seafloor topography, the ocean is not a regular cube for gridding. In addition, the spatio-temporal correlations in various regions are different due to the influence of ocean circulations and climate changes.
(2) \emph{Coupled physical-biogeochemical properties}. The concentration of oceanic dissolved oxygen is influenced by a variety of factors. For one thing, the solubility of oxygen in the water is affected by physical factors, including temperature, salinity and pressure. For another, biological processes, such as photosynthesis and respiration, and organic matter decomposition play a significant role in regulating dissolved oxygen concentrations. 

\textbf{Our Work.} To address the aforementioned challenges, we propose \textsc{OxyGenerator} to perform regression prediction on each 4D coordinate, reconstructing global ocean deoxygenation from 1920 to 2023. Besides, we collect more than 6 billion multi-variable oceanic observation records from multiple databases. Specifically, we first propose graph-based modeling to connect both local and remote observations in irregular four-dimension space. To capture the \emph{spatio-temporal heterogeneity} in different regions, inspired by the zoning strategy in oceanography, we propose zoning-varying graph message-passing via hypernetwork. Moreover, to fuse the knowledge of \emph{physical-biogeochemical properties}, we integrate multiple environmental factors and geographical coordinates for nonlinear feature extraction. Especially for the chemical effects in ocean deoxygenation, we leverage the thermodynamic equilibrium among dissolved oxygen (O), nitrogen (N) and phosphorus (P) to calibrate the uncertainty of reconstruction. 
To summarize, the main contributions of our work are as follows:
\begin{itemize}[left=0.1em,itemsep=3pt,topsep=0pt,parsep=0pt]
    \item To the best of our knowledge, \textsc{OxyGenerator} is the first deep learning based work to reconstruct global ocean deoxygenation over a century from real-world observations, which significantly outperforms the expert-dominated numerical simulation results with a mean absolute percentage error (MAPE) reduction of 38.77\%.
    \item We propose zoning-varying message-passing via graph hypernetwork to obtain consistent 4D spatio-temporal reconstruction, achieving adaptive ocean zoning.
    \item We propose the chemistry-informed gradient variance regularization to calibrate the uncertainty of reconstruction, which combines the inductive bias between dissolved oxygen variation and other nutrients (nitrogen, phosphorus) in chemical equations. 
\end{itemize}
Through our work, we quantitatively calculate that the proportion of oxygen minimum zone where the DO concentration below 30$\mu$mol/kg ($\text{OMZ}_{\text{30}}$) enlarges from 3.54\% in the 1920s to 10.67\% today (Figure \ref{figure1}), alerting us the increasingly urgent issue of global ocean deoxygenation.

\section{Related Work}

In this section, we briefly review the related research lines to our work. A more comprehensive analysis of related work please refer to Appendix \ref{appendix:related_work}.

\textbf{Ocean Deoxygenation.}
Global warming and excessive nutrient inputs caused by human activities have led to significant ocean deoxygenation in recent years. 
To better understand the oxygen cycling mechanism and assess the overall impact of human activities on the marine system since the 20th century, a comprehensive global analysis of ocean deoxygenation is particularly important. Existing research has made two ways of positive attempts, but still leaves some limitations: 
(1) \emph{Numerical Simulation Models}. Existing expert-dominated studies utilize numerical simulations based on climate models to probe the drivers and predict oxygen loss~\cite{RN163, RN165}. For example, Coupled Model Intercomparison Project Phase 6 (CMIP6)~\cite{eyring2016overview} includes three experiments (CESM2-omip1, CESM2-omip2 and GFDL-ESM4-historical) on dissolved oxygen simulation.
Nevertheless, most simulations entirely rely on knowledge of the climate system and fail to leverage observations for correction, thereby showing inferior performance~\cite{RN164, RN166}. 
(2) \emph{Spatial Interpolation Methods}. Due to the severe sparse DO observations, another aspect of studies attempt data reconstruction through distance based weighted average~\cite{Schmidtko2017DeclineIG} and geostatistical regression~\cite{Zhou2022ResponsesOH} for spatial interpolation. However, these methods are only smoothing of the existing DO observation data and yield inaccurate results for areas that lack observation data. They ignore the spatio-temporal heterogeneity of different regions and fail to make full use of auxiliary variables for reconstruction.
In this paper, to the best of our knowledge, we are the first to propose a deep learning method to reconstruct global ocean deoxygenation over a century, considering the spatio-temporal hetegeneity in different regions (Section \ref{sec:4D-st-hypernetwork}) and chemical properties across dissovled oxygen and nutrients (Section \ref{sec:chem-grad-norm}).

\vspace{-1mm}
\textbf{Data-Driven Earth System.}
Existing superior earth system methods are mostly expert-dominated numerical models, which rely on complicated physics processes, sensitive initial conditions and suitable forcing. Moreover, many numerical models are computationally intensive and costly for fine-grained spatio-temporal resolution or long-range simulation. With the rapid growth of artificial intelligence, \emph{AI for Science}, or more specifically data-driven Earth system, has become a hot topic in both Computer Science and Earth Geoscience~\cite{Bergen2019MachineLF,Reichstein2019DeepLA}. 
Given the accumulation of scientific data, data-driven deep learning models are attempting to learn complex and nonlinear correlations in the Earth system, while reducing computational and application costs.
Especially in the field of numerical weather prediction (NWP), deep learning techniques are particularly suitable for improving its prediction performance due to their large amount of data, e.g. recent state-of-the-art methods Pangu-Weather~\cite{bi2023accurate}, GraphCast~\cite{doi:10.1126/science.adi2336}.
For another, ~\citeauthor{Nguyen2023ClimaXAF} propose a foundation model for weather and climate modeling called ClimaX, which extends Transformer architecture for more general weather and climate tasks. Overall, the data-driven Earth system is still in its early stages, and more scenarios and technologies are worth further exploration. 

\begin{figure*}[h]
    \centering
    \includegraphics[width=\linewidth]{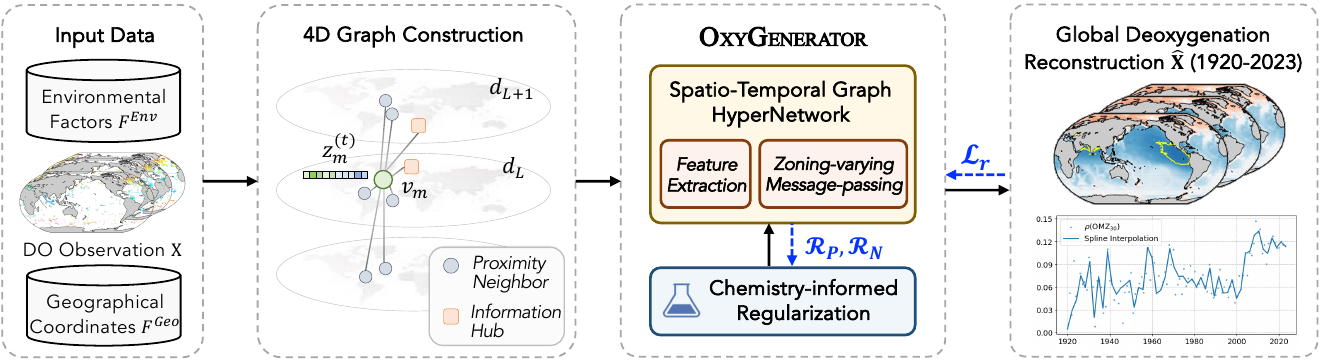}
    \caption{The framework of ocean deoxygenation reconstruction via our proposed \textsc{OxyGenerator}.}
    \label{fig:system_model}
    \vspace{-3mm}
\end{figure*}

\section{Preliminary}
In this section, we first present the problem formulation of reconstructing global ocean deoxygenation, and then introduce the data collection and quality control procedures. Finally, we briefly review the chemical process in ocean deoxygenation. 
Table \ref{tab:symbol} in Appendix \ref{appendix:notation} lists main symbols and notations used throughout this paper. 

\subsection{Problem Formulation}

Reconstruction of global ocean deoxygenation aims to estimate and fill in the missing values within sparse dissolved oxygen observations over the last century. Here we consider the four dimensional coordinate to represent oceanic observation, i.e., longitude, latitude, depth, and time.
Let $\bm{\Omega}=(\omega_{i,j,d,t})_{i,j,d,t}\in \{0,1\}^{L \times G \times D \times T}$be a binary indicator representing observed entries, i.e. $\omega_{i,j,d,t}=1$ \emph{iff} the entry $(i,j,d,t)$ is observed, otherwise it is missing. We denote the incomplete data matrix $\textbf{X}$ as follows:
\begin{equation*}
    \textbf{X} = \textbf{X}^{(obs)} \odot \bm{\Omega} + \text{NA} \odot (\mathbbm{1}_{L \times G \times D \times T} - \bm{\Omega}),
\end{equation*}
where $\textbf{X}^{(obs)} \in \mathbb{R}^{L \times G \times D \times T}$ contains the observed entries, NA denotes the indicator of not available data observation, $\odot$ is the element-wise product and $\mathbbm{1}_{L \times G \times D \times T}$ is an $L \times G \times D \times T$ matrix filled with ones. Given the data matrix $\textbf{X}$, our goal is to construct an estimate $\hat{\textbf{X}}$ filling the missing entries of $\textbf{X}$, which can be written as
\begin{equation*}
    \hat{\textbf{X}} = \textbf{X}^{(obs)} \odot \bm{\Omega} + \hat{\textbf{X}}^{(imp)} \odot (\mathbbm{1}_{L \times G \times D \times T} - \bm{\Omega}),
\end{equation*}
where $\hat{\textbf{X}}^{(imp)}$ contains the imputed values. 
It should be pointed that classical data imputation works (please refer to Section \ref{appendix:related_work} in Appendix) involve manually masking approximately 20\% -80\% of the data to create missing values, and then comparing the imputation results with complete data for performance validation. 
However, due to the high scarcity and unavailability of historical dissolved oxygen records (only 3.735\% of data are observed), we directly conduct regression prediction in each time frame and compare it with the observed DO.
Specifically, we use past $T$ timesteps DO observation $X_{t-T:t-1}$, future $T$ timesteps DO observation $X_{t+1:t+T}$, and some auxiliary variables $\mathcal{D}$ to reconstruct the dissolved oxygen observations at time $t$. Then, we compare them with the actual observations at time $t$ to achieve performance comparison. In other words, we define the following loss function $\mathcal{L}_{r}$ of our model to minimize the reconstruction error:
\begin{equation}
    \label{eq:loss_recons}
    \mathcal{L}_{r} = \sum_t^{} \mathcal{L}(\mathcal{M}(X_{t-T:t-1}, X_{t+1:t+T}, \mathcal{D}) \odot \Omega_t, X_t^{(obs)} \odot \Omega_t),
\end{equation}
where $\mathcal{M}$ is our proposed \textsc{OxyGenerator} parameterized with $\theta$ and $\phi$, $\mathcal{L}$ is the mean squared error loss. 


\subsection{Data Collection and Quality Control}
\label{sec:data-collection}
We collect over 6 billion historical observation records from 1920 to 2023 of dissolved oxygen and relevant environmental factors, including temperature, salinity, nitrates, phosphates, silicates, and chlorophyll, from multiple public databases (Table \ref{tab:data-sources} and Figure \ref{fig:data-collection} in Appendix). We follow the data preprocessing in existing works~\cite{Schmidtko2017DeclineIG} and conduct formatting standardization and spatio-temporal tagging correction for all data. Besides, we establish unified quality control standards to ensure the availability and reliability of observations (Please refer to Appendix \ref{appendix:observ_data} for more information about data quality control). Hereby, we obtain data matrix of dissolved oxygen $\mathbf{X}$, environmental factors $\mathbf{F}^{\text{Env}}$ and geographical coordinates $\mathbf{F}^{\text{Geo}}$. 
In our work, we follow the spatial and temporal resolution setting in \cite{garcia2010world,eyring2016overview}, defined as the annual average temporal resolution from 1920 to 2023, spatial resolution of $1^{\circ} \times 1^{\circ}$, and a total of 33 depth levels from 0 to 5500 meters. Hence, the dimension range of the data is $L=360$, $G=180$, $D=33$ and $T=104$.


\vspace{-2mm}
\subsection{Chemical Effect in Ocean Deoxygenation}
\label{sec:chemical}
Global ocean deoxygenation attributes to complicated factors, among which the chemical effect of oxygen cycling largely impact the dissolved oxygen dynamics.
Specifically, the decomposition of organic matter containing nitrogen (N) and phosphorus (P), and the regeneration of inorganic nutrients hold quantitative relationships with dissolved oxygen variations. \citeauthor{redfield1963influence}(\citeyear{redfield1963influence}) first infer the chemical composition transformation relationship among O, N and P at thermodynamic equilibrium through theoretical analysis as shown in Equation \ref{eq:chemistry}, which is further named Redfield ratio. Afterwards, \citeauthor{ishizu2013relationship} (\citeyear{ishizu2013relationship}) confirm the \emph{linear} correlation of N-O and P-O among above elements through in-situ observation data, which further demonstrates the internal connections of dissolved oxygen variation and chemical effect.
\begin{equation}
\label{eq:chemistry}
\begin{split}
     \underbrace{(\text{CH}_2\text{O})_{106}(\text{NH}_3)_{16}\text{H}_3\text{PO}_4}_{\text{Organic Matter}}+\underbrace{138\text{O}_2}_{\text{DO}}\rightarrow\\106\text{CO}_2+122\text{H}_2\text{O}+\underbrace{16\text{HNO}_3+\text{H}_3\text{PO}_4}_{\text{Inorganic Nutrients}}
\end{split}
\end{equation}
In our work, we combine the domain knowledge in chemical oceanography with our proposed method, leveraging the dynamic equilibrium among environmental factors to improve the reconstruction performance and eliminate abnormal fluctuations. 

\vspace{-2mm}
\section{Methodology}
In this section, we formally introduce the methodology of ocean deoxygenation reconstruction and detailed architecture of our proposed \textsc{OxyGenerator} in Figure \ref{fig:system_model}. 


\vspace{-2mm}
\subsection{Four-Dimensional (4D) Graph Construction}

Ocean observation has three-dimensional spatial coordinates (longitude, latitude and depth), while constrained by land plates and seabed topography, making the entire  data observation an irregular cube.
Meanwhile, due to the sparsity of observations, when there is a lack of observation information in adjacent receptive fields, localized convolution filters cannot  aggregate sufficient information. 
Therefore, instead of CNN-based methods, we propose graph modeling to define two types of neighbors, i.e., \emph{proximity neighbor} and \emph{information hub}, for each data grid. We consider 3D spatial proximity under irregular boundaries. At the same time, we connect a wider range of related nodes with respect to the observation completeness, hereby improving the richness of information.



\textbf{Proximity Neighbor.} The \emph{First Law of Geography}~\cite{tobler1970computer} states that "everything is related to everything else, but near things are more related than distant things." 
According to that, we consider proximity neighbors as data grids within the range of $[-\delta^{\circ}, \delta^{\circ}]$ in the horizontal longitude and latitude range, as well as ±$d_m$ depth layer in the vertical direction. Regarding the irregular oceanic boundaries, we adopt bedrock elevation data published by NOAA to ensure the rationality.

\textbf{Information Hub.} Due to the insufficient observations in proximity neighbors, we further expand the spatial proximity range and utilize observation completeness $\mathcal{C}_{\text{obs}}$ as an indicator to measure the richness of information within a time range from $t-T$ to $t+T$:
\begin{equation*}
    \mathcal{C}_{\text{obs}} = \frac{\left\|\omega_{i,j,d,t-T:t+T} \right\|_1}{2T} = \frac{\sum_{\tau} \left\|\omega_{i,j,d,\tau}\right\|}{2T}.
\end{equation*}
When $\mathcal{C}_{\text{obs}} \ge \varepsilon_{\mathcal{C}}$, we deem this data grid an information hub and the neighbor of node $v_{i,j,d,t}$, otherwise it is not.
Thanks to the flexibility of graph modeling, information hub can effectively enlarge the receptive field, which is crucial for reconstruction under sparse observations.
We define the adjacency matrix $A_t$ and node set $\mathcal{V}_t$ at time $t$,  and the corresponding edge feature matrix $\mathcal{E}_t$ as the difference of attributes, including distance, density, pressure, etc. 

\vspace{-2mm}
\subsection{Spatio-Temporal Graph HyperNetwork}
\label{sec:4D-st-hypernetwork}

\textbf{Feature Extractor.} The concentration changes of dissolved oxygen are influenced by different factors. Thus, we first want to construct a feature extractor $f(\theta)$ that can fully characterize the attribute features of nodes. In order to capture the temporal variation of dissolved oxygen, we adopt a bidirectional LSTM model $f_{\text{Bi-LSTM}}$ to encode both historical and future $T$ timesteps DO observation as follows:
\begin{align*}
    \vspace{-2mm}
    \overrightarrow{Z_\tau^{\text{DO}}} &= \text{Bi-LSTM}(X_{\tau}, \overrightarrow{Z_{\tau-1}^{\text{DO}}}; \overrightarrow{\theta}_{\text{Bi-LSTM}}),\\
    \overleftarrow{Z_\tau^{\text{DO}}} &= \text{Bi-LSTM}(X_{\tau}, \overleftarrow{Z_{\tau-1}^{\text{DO}}}; \overleftarrow{\theta}_{\text{Bi-LSTM}}),
    \vspace{-2mm}
\end{align*}
where $X_\tau$ denotes the DO observation at time $\tau$, $\overrightarrow{Z_\tau^{\text{DO}}}$ and $\overleftarrow{Z_\tau^{\text{DO}}}$ denote the hidden states from two directions, $\overrightarrow{\theta}_{\text{Bi-LSTM}}$ and $\overleftarrow{\theta}_{\text{Bi-LSTM}}$ denote the Bi-LSTM parameters. Correspondingly, the DO temporal feature at reconstruction time $t$ is $Z_t^{\text{DO}} = \overrightarrow{Z_t^{\text{DO}}} \| \overleftarrow{Z_t^{\text{DO}}}$, thereby capturing the temporal evolution of DO in two directions. In addition, dissolved oxygen concentration is also related to a series of geographical coordinates $F_t^{\text{Geo}}$ (including latitude, longitude, depth, time) and environmental factors $F_t^{\text{Env}}$ (including temperature, salinity, nutrients, chlorophyll). 
In order to comprehensively consider the coupling correlations between these factors and dissolved oxygen, we leverage a multi-layer perceptron (MLP) to embed into a latent space: 
$Z_{t}^{F} = \text{MLP}(F_t^{\text{Geo}} \| F_t^{\text{Env}}; \theta_{MLP})$. To sum up, the overall latent feature embedding is the composition of both DO temporal feature $Z_t^{\text{DO}}$ and multi-factor feature $Z_{t}^{F}$ as $Z_t = Z_t^{\text{DO}} \| Z_{t}^{F}$.

\begin{figure}
    \centering
    \includegraphics[width=\linewidth]{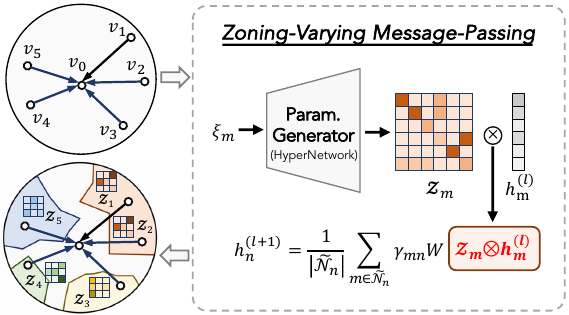}
    \caption{Framework of zoning-varying message-passing.}
    \label{fig:zoning-varying}
    \vspace{-4mm}
\end{figure}

\textbf{Zoning-Varying Message-Passing.}
The global ocean over a century shows heterogeneous spatio-temporal correlations in different historical periods and regions due to the varying impacts of human activities and climate change. In oceanographic research, many studies partition the global ocean into different zones~\cite{fay2014global,Reygondeau2020ClimateCE,Sonnewald2020ElucidatingEC} in order to better investigate the complex oceanic processes. Inspired by zoning strategy in oceanography, a naive method is to train multiple models through pre-determined zones. However, there have no theoretical basis for the partition of deoxygenation areas, and training multiple models increase the computational costs. Therefore, we propose \textit{Spatio-Temporal Graph HyperNetwork}, which adaptively generate zoning-varying parameters via hypernetwork~\cite{DBLP:conf/iclr/HaDL17} for graph message-passing. 
We can obtain a dynamic partition more efficiently through a globally shared parameter generator and the low dimensional context information of each node.

To simplify the explanation, we denote the $m$-th node in the node set $\mathcal{V}_t = \{ v_{i,j,d,t}\}_{i,j,d,t}$ at time $t$ as $v_m$.
For each node $v_m$, we generate zoning-varying parameter $\mathcal{Z}_m = [\mathcal{Z}^{\alpha}_m, \mathcal{Z}_m^{\beta}]$ based on context information $\xi_m$ which includes geographical coordinates and physical elements:
\begin{equation*}
    \mathcal{Z}^{\alpha}_m = \text{MLP}_\alpha(\xi_m; \phi_\alpha), \mathcal{Z}_m^{\beta} = \text{MLP}_\beta(\xi_m; \phi_\beta),
\end{equation*}
where $\mathcal{Z}^{\alpha}_m \in \mathbb{R}^{d_z \times d_z}$ denotes rotation matrix, $\mathcal{Z}_m^{\beta} \in \mathbb{R}^{d_z}$ denotes bias vector. Afterwards, compared to vanilla graph neural networks (GNN), we perform non-shared feature scaling on the latent feature embedding:
\begin{equation*}
    \tilde{h}_m^{(l)} = h_m^{(l)} \otimes \mathcal{Z} \triangleq {\mathcal{Z}^{\alpha}_m}^T \cdot h_m^{(l)} + \mathcal{Z}_m^{\beta},
\end{equation*}
where $h_m^{(l)}$ is the $l$-th layer GNN input of node $v_m$, $\tilde{h}_m^{(l)}$ is the node-wise adaptive latent feature of $h_m^{(l)}$.
Thus, the nonlinear correlation between different geographic information and physical elements will serve as a criterion for dynamic partitioning. 
Nodes with similar zoning parameter $\mathcal{Z}$ form the same zone.
Unlike unsupervised clustering or manually setting partition rules, for ocean deoxygenation problems, hypernetworks can transform discrete partitions into continuous ones. We further use graph neural networks for message passing on adaptive latent features and consider the edge features of different neighbors. Take node $v_n$ as an example, the $l+1$-th layer GNN output is denoted as:
\begin{equation*}
    h_n^{(l+1)} = \frac{1}{|\tilde{\mathcal{N}}_n|}\sum\nolimits_{m\in \tilde{\mathcal{N}}_n} \gamma_{mn} W \tilde{h}_m^{(l)},
\end{equation*}
where $\tilde{z}_{m}^{l}$ is the adaptive latent feature, $W$ is the model parameter of message-passing, and $\tilde{\mathcal{N}}_n$ is the neighbor set of node $v_n$ with self-loop.
The edge weight $\gamma_{mn}$ is derived by nonlinear projection of edge feature by a multi-layer perceptron: $\gamma_{mn} = \text{MLP}_{\gamma}(e_{mn})$.

\vspace{-2mm}
\subsection{Chemistry-informed Regularization}
\label{sec:chem-grad-norm}
The global ocean deoxygenation is a complex system that couples physical and biochemical effects. 
The embedding of domain knowledge can calibrate the uncertainty of neural networks and eliminate abnormal reconstruction. 
However, despite some research on physics-informed neural networks~\cite{DBLP:journals/jcphy/RaissiPK19,DBLP:conf/icml/PodinaEK23,DBLP:conf/iclr/ZengKBS23}, there is still little research on how to express chemical dynamic equilibrium as shown in Equation \ref{eq:chemistry}. 
Here, we consider the dynamic transition equilibrium between dissolved oxygen and nitrate (phosphate), which means the gradient between dissolved oxygen and nitrate (phosphate) concentration is a constant. For example, for the observation of nitrate $F_m^{N}$ (phosphate $F_m^{P}$) at node $v_m$ (node $v_n$), we can calculate the corresponding gradient based on the reconstructed dissolved oxygen $\hat{x}_m$ ($\hat{x}_n$) which is approximately a constant, i.e., $\frac{\partial \hat{x}_m}{\partial F_m^{N}} = \text{const.}$, $\frac{\partial \hat{x}_n}{\partial F_n^{P}} = \text{const.}$

Therefore, we propose the chemistry-informed gradient variance regularization as one of the supervised signals. Specifically, in a batch of training data, we select nodes with nitrate (phosphate) observations and calculate their corresponding gradients. Since the dynamic equilibrium between nitrate (phosphate) and dissolved oxygen is approximately consistent, the norm of gradient variance should be small:
\begin{equation}
    \label{eq:loss_chem}
    \mathcal{R}_{N} = \left\|\sigma \left( \left\{ \frac{\partial \hat{x}_m}{\partial F_m^{N}} \right\}_m \right)\right\|_2^2, \,
    \mathcal{R}_{P} = \left\|\sigma \left( \left\{ \frac{\partial \hat{x}_n}{\partial F_n^{P}} \right\}_n\right)\right\|_2^2,
\end{equation}
where $\sigma(\cdot)$ denotes the calculation of the gradient variance.

\subsection{Optimization Algorithm}
Due to the memory burden of large scale graphs for global ocean, we divide it into different training groups based on World Ocean Database 2018~\cite{wod2018} ocean division. Since different groups show varying complexity, each group is optimized using different iterations in one epoch of training. We periodically evaluate the reconstruction performance of different group using the validation dataset, and then reset the iterations number for each group. During the learning process, we integrate the reconstruction loss $\mathcal{L}_r$ in Equation \ref{eq:loss_recons} and two chemical-informed gradient variance regularization in Equation \ref{eq:loss_chem} as follows:
\begin{equation}
    \label{eq:loss_oxy}
    \mathcal{L}_{\textsc{OxyGenerator}} = \mathcal{L}_{r} + \lambda (\mathcal{R}_{N} + \mathcal{R}_{P}),
\end{equation}
where $\lambda$ is the ratio coefficient of two losses. We conclude the algorithm in Appendix \ref{appendix:algorithm} Algorithm \ref{alg:oxy}.

\vspace{-2mm}

\section{Experiment}

In this section, we evaluate the effectiveness of our proposed \textsc{OxyGenerator} in reconstructing global ocean deoxygenation from 1920 to 2023. More comprehensive in-depth analysis are presented in detail with the aim of answering the following four research questions:
\begin{itemize}[left=0.1em,itemsep=3pt,topsep=0pt,parsep=0pt]
    \item \textbf{RQ1}: Can the \textsc{OxyGenerator} reconstruct the global deoxygenation trend? How does its performance compare with expert-dominated numerical simulation methods?
    \item \textbf{RQ2}: What are the learning pattern and insights of \textsc{OxyGenerator} during training?
    \item \textbf{RQ3}: How effective are different parts of the model?
    \item \textbf{RQ4}: What are the shortcomings of deep learning based model \textsc{OxyGenerator} and future directions?
\end{itemize}

\begin{table*}
\centering
\caption{Comparison with simulation results from CMIP6. The best results are highlighted in bold, and the second best is underlined. We calculate the performance improvement of \textsc{OxyGenerator} compared to the suboptimal one.}
\label{tab:performance-comparison}
\resizebox{\linewidth}{!}{%
\begin{tabular}{@{}lcccccccc@{}}
\toprule
\multirow{2}{*}{Benchmark} & $k=1$ & $k=2$ & $k=3$ & $k=4$ & \multicolumn{4}{c}{Average Performance} \\ \cmidrule(l){2-9} 
 & MAPE  & MAPE  & MAPE  & MAPE  & MAPE  & R2 & RMSE & MAE \\ \midrule
CESM2 omip1 & 23.63 & 23.15 & \underline{23.67} & \underline{22.87} & \underline{23.32\scriptsize{±0.38}} & 0.7966\scriptsize{±0.0064} & 37.37\scriptsize{±0.34} & 25.98\scriptsize{±0.31} \\
CESM2 omip2 & \underline{23.62} & \underline{23.00} & 24.60 & 23.13 & 23.58\scriptsize{±0.72} & 0.7947\scriptsize{±0.0096} & 38.22\scriptsize{±0.55} & 27.12\scriptsize{±0.32} \\
GFDL-ESM4 historical & 26.13 & 24.01 & 26.68 & 24.33 & 25.28\scriptsize{±1.31} & \underline{0.8228\scriptsize{±0.0051}} & \underline{35.45\scriptsize{±0.65}} & \underline{23.69\scriptsize{±0.38}} \\ \midrule
\textsc{OxyGenerator} (Ours) & \textbf{14.72} & \textbf{13.48} & \textbf{15.72} & \textbf{13.20} & \textbf{14.28\scriptsize{±1.16}} & \textbf{0.9026\scriptsize{±0.0072}} & \textbf{26.31\scriptsize{±1.23}} & \textbf{17.57\scriptsize{±1.10}} \\
Improvement & 37.67\% & 41.38\% & 33.59\% & 42.28\% & 38.77\% & 9.70\% & 25.78\% & 25.83\% \\ \bottomrule
\end{tabular}%
}
\end{table*}

\begin{figure*}
    \vspace{-2mm}
    \centering
    \includegraphics[width=\linewidth]{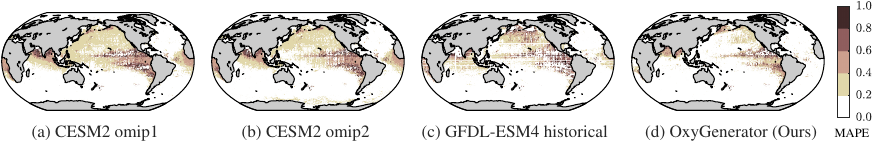}
    \caption{The spatial distribution of MAPE in global ocean deoxygenation reconstruction using different methods (The darker the color, the greater the error). The reconstruction error of \textsc{OxyGenerator} in open sea is significantly reduced.}
    \label{fig:exp-global-mape}
\end{figure*}

\textbf{Evaluation Methods.}
Due to the irreproducibility of historical ocean observations, only existing observations can be utilized for evaluation. Meanwhile, due to the data scarcity, further data partitioning for training/validation/testing would render the interpolation and imputation algorithms inapplicable.
Therefore, we treat the deoxygenation reconstruction as a regression task  independent of current-time observations and perform 4-fold cross testing of the collected data. For each fold, we randomly choose 25\% observation data as the test data and the rest as training and validation. We report on the performance on each fold test data and the average performance on 4 folds.

\textbf{Evaluation Metrics.}
We employ four metrics for performance evaluation, including Root Mean Square Error (RMSE), Mean Absolute Percentage Error (MAPE), Mean Absolute Error (MAE), and Coefficient of Determination ($R^2$). More details please refer to Appendix \ref{appendix:exp-details}.

\textbf{Gold-standard Benchmark.}
As we are the first research to use deep learning methods for global dissolved oxygen reconstruction over a century, our method is benchmarked against three advanced expert-dominated simulation results from Coupled Model Inter-comparison Project Phase 6 (CMIP6): \textbf{CESM2 omip1}~\cite{CESM2}, \textbf{CESM2 omip2}~\cite{CESM2} and \textbf{GFDL-ESM4 historical}~\cite{GFDL-ESM4}. 
The details of three simulation experiment are introduced in Appendix \ref{appendix:cmip6}.

\begin{figure}
    \centering
    \includegraphics[width=\linewidth]{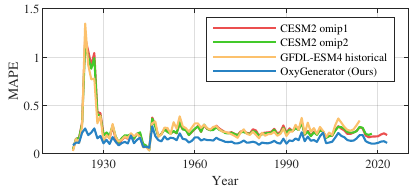}
    \caption{Performance Comparison of MAPE over time.}
    \label{fig:exp-time-mape}
    \vspace{-5mm}
\end{figure}

\vspace{-2mm}
\subsection{Experiment Results (RQ1)}

In Table \ref{tab:performance-comparison}, we illustrate the comparison results of 4-fold cross testing.
Our \textsc{OxyGenerator} achieves the best performance on all four metrics, with a 38.77\% reduction in MAPE compared to suboptimal numerical simulation methods. It proves that deep learning methods can more accurately reconstruct ocean deoxygenation trends. 

\textbf{Horizontal Spatial Distribution of Reconstruction Error.} Figure \ref{fig:exp-global-mape} depicts the average MAPE of four methods in spatial distribution. Owing to the strength of capturing heterogeneous spatio-temporal correlations, \textsc{OxyGenerator} provides more accurate reconstruction in the North Pacific, Indian Ocean, Equatorial Atlantic and other regions.

\textbf{Temporal Distribution of Reconstruction Error.} We analyze the variation of reconstruction error over time, as shown in Figure \ref{fig:exp-time-mape}. 
\textsc{OxyGenerator} shows the consistent reconstruction performance among different years.
Notably, there exist an error peak in 1923-1928. This large estimation error in three simulation models comes from a severe underestimation of the Black Sea deoxygenation (Detailed analysis on the Black Sea can be found in Appendix \ref{appendix:black-sea}). \textsc{OxyGenerator}, based on the correction and correlation learning of observations, effectively captures the changes in extreme deoxygenation areas.

\textbf{Vertical Depth Distribution of Reconstruction Error.} We compare variation of DO reconstruction within the depths of 0-5500 meters for the five oceans (i.e., Pacific, Atlantic, Indian, Antarctic and Arctic) in Appendix \ref{appendix:vertical}.

\textbf{Comparison with Other Deep Learning Models.} In addition to expert-dominated numerical simulation, \textsc{OxyGenerator} outperforms other deep learning models such as MLP, GNN, Transformer, etc., as shown in Appendix \ref{appendix:deep-learning-model}.

\subsection{Model Analysis (RQ2)}

\begin{figure}
    \centering
    \includegraphics[width=\linewidth]{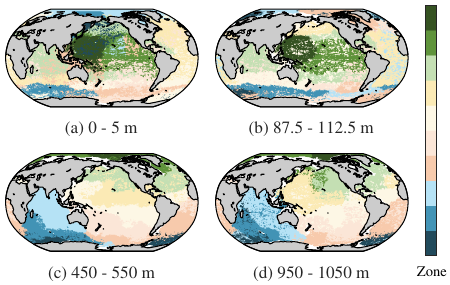}
    \caption{Adaptive zoning of 4 representative depth levels in 2016. Different colors represent different zones.}
    \label{fig:hypernetwork}
    \vspace{-4mm}
\end{figure}

\textbf{Effect of Adpative Zoning via Spatio-Temporal HyperNetwork.} \label{Adaptive_zoningl}
We randomly select four representative depth levels in 2016 and visualize the generated parameters $\mathcal{Z}$ through T-SNE.
As shown in Figure \ref{fig:hypernetwork}, \textsc{OxyGenerator} can adaptively carry out spatial zoning, which is quantified into 10 zones. 
Moreover, compared with WOD zoning strategy~\cite{wod2018}, we observe that:
(1) The adaptive zoning is closely related to latitude distribution and presents a clear oceanic partitioning. Especially below 450 meters depth, we show similar zoning results with WOD in the Indian Ocean, Atlantic Ocean, and Pacific Ocean. (2) Compared to depth-independent zoning in WOD, our method depicts a finer grained zoning that varies at different depths and intra-ocean. Surface seawater is influenced more by human activities and exhibits a more intricate zoning pattern, whereas deep seawater is closely associated with geographical location. More details refer to Appendix \ref{appendix:zoning}.

\begin{figure}
    \centering
    \includegraphics[width=\linewidth]{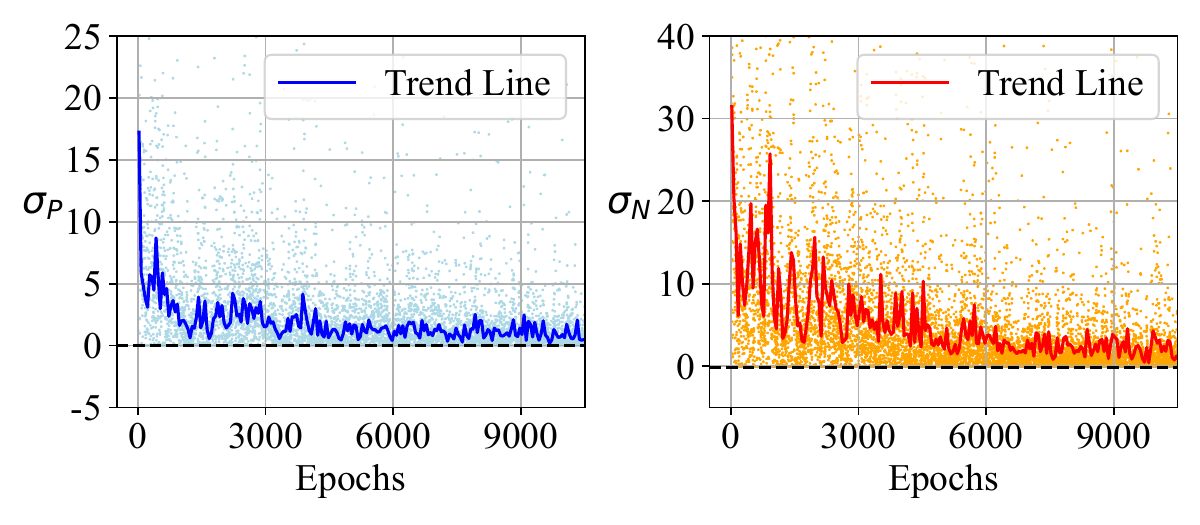}
    \caption{The gradient changes of $O-P$ and $O-N$ during the training process. Each point represents the average gradient value and two lines are corresponding trend lines.}
    \label{fig:exp-gradient}
    \vspace{-4mm}
\end{figure}

\textbf{Analysis of Chemistry-informed Gradient Variation During Training.} Figure \ref{fig:exp-gradient} shows the variance changes of two gradient during training, i.e. $\sigma_P = \sigma(\frac{\partial O}{\partial P})$ and $\sigma_N = \sigma(\frac{\partial O}{\partial N})$. According to the chemical equation \ref{eq:chemistry}, the gradients of two element are equal to two constant of $-\frac{1}{138}$ and $-\frac{16}{138}$. 
The variance of two gradient gradually approaches to zero during the training process, which effectively reflects the thermodynamic equilibrium among dissolved oxygen (O), nitrogen (N) and phosphorus (P).

\subsection{Ablation Study (RQ3)}
To verify the effectiveness of our proposed \textsc{OxyGenerator}, we conduct five degenerate variants for ablation study. As shown in Table \ref{tab:performance-ablation}, any variant exhibits performance degradation, indicating the effectiveness of our proposed method. Especially when we remove the zoning-varying graph message-passing, the reconstruction error significantly increase due to the failure of capturing spatio-temporal heterogeneity over long-range time and space.

\begin{table}[]
\centering
\caption{Ablation Study of \textsc{OxyGenerator}.}
\label{tab:performance-ablation}
\renewcommand{\arraystretch}{1.15}
\resizebox{\linewidth}{!}{%
\begin{tabular}{lccc}
\hline
Ablation Variants & MAPE & RMSE & MAE \\ \hline
vanilla MLP & 23.03\scriptsize{±1.29} & 30.89\scriptsize{±0.95} & 22.08\scriptsize{±1.01} \\
w/o Zoning-Varying & 23.84\scriptsize{±0.88}  & 30.97\scriptsize{±0.90}   & 22.25\scriptsize{±1.24} \\
w/o Chem. Regularization & 14.85\scriptsize{±1.41} & 26.68\scriptsize{±0.81} & 18.06\scriptsize{±0.50} \\
w/o Enviromental Factors & 15.70\scriptsize{±1.42} & 27.95\scriptsize{±0.61} &18.94\scriptsize{±0.70}  \\ 
w/o Temporal DO Obs. &17.01\scriptsize{±1.04}  &29.26\scriptsize{±0.91}  & 20.39\scriptsize{±0.66} \\ \hline
\textsc{OxyGenerator} (Ours) & \textbf{14.28\scriptsize{±1.16}} & \textbf{26.31\scriptsize{±1.23}} & \textbf{17.57\scriptsize{±1.10}} \\ \hline
\end{tabular}%
}
\vspace{-3mm}
\end{table}

\subsection{Limitations and Future Directions (RQ4)}
\begin{figure}
    \centering
    \includegraphics[width=\linewidth]{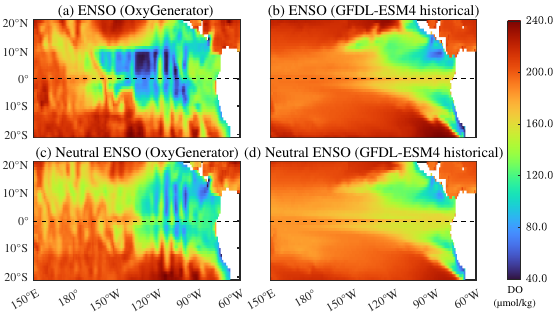}
    \caption{Comparison of reconstruction results in the surface of the Tropical Pacific during ENSO and neutral ENSO.}
    \label{fig:exp-enso}
    \vspace{-3mm}
\end{figure}

Although \textsc{OxyGenerator} achieve promising performance, we discover some limitations and future directions. (1) Due to the scarcity of observational data, there is significant uncertainty in evaluating the effectiveness of different methods.
For example, the South Pacific accounts for only 0.027\% of complete observations in 1920-1950. Therefore, discovering more data from databases or literature is an important proposition in the era of open science data~\cite{10.1145/3290605.3300356,10.1145/3543873.3587305}. (2) The reconstruction performance requires validation according to the laws of oceanography.
Figure \ref{fig:exp-enso} shows the reconstruction results of the tropical Pacific region under El Niño-Southern Oscillation (ENSO) and neutral ENSO. 
In El Niño years, with the increase of seawater temperature in the tropical eastern Pacific, the concentration of dissolved oxygen will decrease, which is clearly reflected in our reconstruction results, but not in numerical simulation. 
However, the extent and impact of the DO reduction require further analysis with oceanographers.


\vspace{-2mm}
\section{Conclusion}
In this paper, we propose \textsc{OxyGenerator}, a deep learning model that effectively reconstructs global ocean deoxygenation based on sparse observation data in 1920-2023. \textsc{OxyGenerator} improves the accuracy of the reconstruction through zoning-varying graph message-passing and chemistry-informed regularization, with a significant improvement compared to expert-dominated numerical simulations. In the future, we will continue to collaborate with oceanographers to further improve the compliance with physical-biogeochemical mechanisms and investigate its impacts for marine ecosystem.


\section*{Acknowledgements}

This work was partially supported by National Natural Science Foundation of China (No. 62020106005, 62272301, 623B2071), National Key R\&D Program of China (No.2022YFB3904204), National Natural Science Foundation of China (No. 42125601, 42276201, 42050105, 62061146002), and Shanghai Pilot Program for Basic Research - Shanghai Jiao Tong University.

\section*{Impact Statements}

This paper contributes to reconstruct the global ocean deoxygenation over a century with deep learning, quantitatively assessing the trends of "breathless ocean". The potential broader impact of this work is notable as follows:
\begin{itemize}[left=0.1em,itemsep=3pt,topsep=0pt,parsep=0pt]
    \item \textbf{Global Ocean Health}: By leveraging AI for comprehensive deoxygenation reconstruction, our study enhances our understanding of historical patterns and contemporary trends of dissolved oxygen variation. The quantitative assessment of oceanic oxygen deficiency aids in identifying regions susceptible to deoxygenation, enabling targeted conservation efforts. 
    \item \textbf{Climate Change Awareness}: Our research provides data support for analyzing the complex relationship between ocean oxygen levels and climate change. We can help improve our understanding of climate change and emphasize the interconnection between ocean health and broader environmental challenges.
    \item \textbf{Interdisciplinary Scientific Collaboration}: The methodologies presented in this paper foster collaboration between the fields of marine science and artificial intelligence. By encouraging interdisciplinary research, we promote the integration of advanced technologies into oceanography, paving the way for innovative solutions to global challenges. Meanwhile, oceanography has also provided inspiration and new demands for AI technology.
\end{itemize}

\bibliography{icml2023/reference}

\begin{thebibliography}{57}
\providecommand{\natexlab}[1]{#1}
\providecommand{\url}[1]{\texttt{#1}}
\expandafter\ifx\csname urlstyle\endcsname\relax
  \providecommand{\doi}[1]{doi: #1}\else
  \providecommand{\doi}{doi: \begingroup \urlstyle{rm}\Url}\fi

\bibitem[Adhikari et~al.(2023)Adhikari, Jiang, Zhan, He, Rawat, Aickelin, and Khorshidi]{DBLP:journals/csur/AdhikariJZHRAK23}
Adhikari, D., Jiang, W., Zhan, J., He, Z., Rawat, D.~B., Aickelin, U., and Khorshidi, H.~A.
\newblock A comprehensive survey on imputation of missing data in internet of things.
\newblock \emph{{ACM} Comput. Surv.}, 55\penalty0 (7):\penalty0 133:1--133:38, 2023.

\bibitem[Bergen et~al.(2019)Bergen, Johnson, de~Hoop, and Beroza]{Bergen2019MachineLF}
Bergen, K.~J., Johnson, P.~A., de~Hoop, M.~V., and Beroza, G.~C.
\newblock Machine learning for data-driven discovery in solid earth geoscience.
\newblock \emph{Science}, 363, 2019.

\bibitem[Bi et~al.(2023)Bi, Xie, Zhang, Chen, Gu, and Tian]{bi2023accurate}
Bi, K., Xie, L., Zhang, H., Chen, X., Gu, X., and Tian, Q.
\newblock Accurate medium-range global weather forecasting with 3d neural networks.
\newblock \emph{Nature}, 619\penalty0 (7970):\penalty0 533--538, 2023.

\bibitem[Bopp et~al.(2013)Bopp, Resplandy, Orr, Doney, Dunne, Gehlen, Halloran, Heinze, Ilyina, Séférian, Tjiputra, and Vichi]{RN166}
Bopp, L., Resplandy, L., Orr, J.~C., Doney, S.~C., Dunne, J.~P., Gehlen, M., Halloran, P., Heinze, C., Ilyina, T., Séférian, R., Tjiputra, J., and Vichi, M.
\newblock Multiple stressors of ocean ecosystems in the 21st century: projections with cmip5 models.
\newblock \emph{Biogeosciences}, 10\penalty0 (10):\penalty0 6225--6245, 2013.

\bibitem[Bopp et~al.(2017)Bopp, Resplandy, Untersee, Le~Mezo, and Kageyama]{RN165}
Bopp, L., Resplandy, L., Untersee, A., Le~Mezo, P., and Kageyama, M.
\newblock Ocean (de)oxygenation from the last glacial maximum to the twenty-first century: insights from earth system models.
\newblock \emph{Philosophical Transactions of the Royal Society A: Mathematical, Physical and Engineering Sciences}, 375\penalty0 (2102):\penalty0 20160323, 2017.

\bibitem[Boyer et~al.(2018)Boyer, Baranova, Coleman, Garcia, Grodsky, Locarnini, Mishonov, Paver, Reagan, Seidov, Smolyar, Weathers, and Zweng]{wod2018}
Boyer, T., Baranova, O., Coleman, C., Garcia, H., Grodsky, A., Locarnini, R., Mishonov, A., Paver, C., Reagan, J., Seidov, D., Smolyar, I., Weathers, K., and Zweng, M.
\newblock World ocean database 2018.
\newblock Technical Report~87, NOAA Atlas NESDIS, 2018.
\newblock \url{https://www.ncei.noaa.gov/sites/default/files/2020-04/wod_intro_0.pdf}.

\bibitem[Breitburg et~al.(2018)Breitburg, Levin, Oschlies, Grégoire, Chavez, Conley, Garçon, Gilbert, Gutiérrez, Isensee, Jacinto, Limburg, Montes, Naqvi, Pitcher, Rabalais, Roman, Rose, Seibel, Telszewski, Yasuhara, and Zhang]{doi:10.1126/science.aam7240}
Breitburg, D., Levin, L.~A., Oschlies, A., Grégoire, M., Chavez, F.~P., Conley, D.~J., Garçon, V., Gilbert, D., Gutiérrez, D., Isensee, K., Jacinto, G.~S., Limburg, K.~E., Montes, I., Naqvi, S. W.~A., Pitcher, G.~C., Rabalais, N.~N., Roman, M.~R., Rose, K.~A., Seibel, B.~A., Telszewski, M., Yasuhara, M., and Zhang, J.
\newblock Declining oxygen in the global ocean and coastal waters.
\newblock \emph{Science}, 359\penalty0 (6371):\penalty0 eaam7240, 2018.

\bibitem[Cao et~al.(2018)Cao, Wang, Li, Zhou, Li, and Li]{DBLP:conf/nips/CaoWLZLL18}
Cao, W., Wang, D., Li, J., Zhou, H., Li, L., and Li, Y.
\newblock {BRITS:} bidirectional recurrent imputation for time series.
\newblock In \emph{Advances in Neural Information Processing Systems 31: Annual Conference on Neural Information Processing Systems 2018, NeurIPS 2018, December 3-8, 2018, Montr{\'{e}}al, Canada}, pp.\  6776--6786, 2018.

\bibitem[Cheung et~al.(2013)Cheung, Sarmiento, Dunne, Fr{\"o}licher, Lam, Palomares, Watson, and Pauly]{Cheung2013ShrinkingOF}
Cheung, W. W.~L., Sarmiento, J.~L., Dunne, J.~P., Fr{\"o}licher, T.~L., Lam, V. W.~Y., Palomares, M. L.~D., Watson, R.~A., and Pauly, D.
\newblock Shrinking of fishes exacerbates impacts of global ocean changes on marine ecosystems.
\newblock \emph{Nature Climate Change}, 3:\penalty0 254--258, 2013.

\bibitem[Cini et~al.(2022)Cini, Marisca, and Alippi]{DBLP:conf/iclr/CiniMA22}
Cini, A., Marisca, I., and Alippi, C.
\newblock Filling the g{\_}ap{\_}s: Multivariate time series imputation by graph neural networks.
\newblock In \emph{The Tenth International Conference on Learning Representations, {ICLR} 2022, Virtual Event, April 25-29, 2022}, 2022.

\bibitem[Claret et~al.(2018)Claret, Galbraith, Palter, Bianchi, Fennel, Gilbert, and Dunne]{RN162}
Claret, M., Galbraith, E.~D., Palter, J.~B., Bianchi, D., Fennel, K., Gilbert, D., and Dunne, J.~P.
\newblock Rapid coastal deoxygenation due to ocean circulation shift in the northwest atlantic.
\newblock \emph{Nature Climate Change}, 8\penalty0 (10):\penalty0 868--872, 2018.

\bibitem[Danabasoglu et~al.(2020)Danabasoglu, Lamarque, Bacmeister, Bailey, A, Edwards, Emmons, Fasullo, Garcia, Gettelman, Hannay, Holland, Large, Lauritzen, Lawrence, Lenaerts, Lindsay, Lipscomb, Mills, Neale, Oleson, Otto‐Bliesner, Phillips, Sacks, Tilmes, Van~Kampenhout, Vertenstein, Bertini, Dennis, Deser, Fischer, Fox‐Kemper, Kay, Kinnison, Kushner, Larson, Long, Mickelson, Moore, Nienhouse, Polvani, Rasch, and Strand]{CESM2}
Danabasoglu, G., Lamarque, J.~F., Bacmeister, J., Bailey, D.~A., A, K., Edwards, J., Emmons, L.~K., Fasullo, J., Garcia, R., Gettelman, A., Hannay, C., Holland, M.~M., Large, W.~G., Lauritzen, P.~H., Lawrence, D.~M., Lenaerts, J. T.~M., Lindsay, K., Lipscomb, W.~H., Mills, M.~J., Neale, R., Oleson, K.~W., Otto‐Bliesner, B., Phillips, A.~S., Sacks, W., Tilmes, S., Van~Kampenhout, L., Vertenstein, M., Bertini, A., Dennis, J., Deser, C., Fischer, C., Fox‐Kemper, B., Kay, J.~E., Kinnison, D., Kushner, P.~J., Larson, V.~E., Long, M.~C., Mickelson, S., Moore, J.~K., Nienhouse, E., Polvani, L., Rasch, P.~J., and Strand, W.~G.
\newblock The community earth system model version 2 (cesm2).
\newblock \emph{Journal of Advances in Modeling Earth Systems}, 12\penalty0 (2), 2020.
\newblock ISSN 1942-2466.
\newblock \doi{10.1029/2019ms001916}.
\newblock URL \url{https://dx.doi.org/10.1029/2019MS001916}.

\bibitem[Dunne et~al.(2020)Dunne, Horowitz, Adcroft, Ginoux, Held, John, Krasting, Malyshev, Naik, Paulot, Shevliakova, Stock, Zadeh, Balaji, Blanton, Dunne, Dupuis, Durachta, Dussin, Gauthier, Griffies, Guo, Hallberg, Harrison, He, Hurlin, McHugh, Menzel, Milly, Nikonov, Paynter, Ploshay, Radhakrishnan, Rand, Reichl, Robinson, Schwarzkopf, Sentman, Underwood, Vahlenkamp, Winton, Wittenberg, Wyman, Zeng, and Zhao]{GFDL-ESM4}
Dunne, J.~P., Horowitz, L.~W., Adcroft, A.~J., Ginoux, P., Held, I.~M., John, J.~G., Krasting, J.~P., Malyshev, S., Naik, V., Paulot, F., Shevliakova, E., Stock, C.~A., Zadeh, N., Balaji, V., Blanton, C., Dunne, K.~A., Dupuis, C., Durachta, J., Dussin, R., Gauthier, P. P.~G., Griffies, S.~M., Guo, H., Hallberg, R.~W., Harrison, M., He, J., Hurlin, W., McHugh, C., Menzel, R., Milly, P. C.~D., Nikonov, S., Paynter, D.~J., Ploshay, J., Radhakrishnan, A., Rand, K., Reichl, B.~G., Robinson, T., Schwarzkopf, D.~M., Sentman, L.~T., Underwood, S., Vahlenkamp, H., Winton, M., Wittenberg, A.~T., Wyman, B., Zeng, Y., and Zhao, M.
\newblock The gfdl earth system model version 4.1 (gfdl‐esm 4.1): Overall coupled model description and simulation characteristics.
\newblock \emph{Journal of Advances in Modeling Earth Systems}, 12\penalty0 (11), 2020.
\newblock ISSN 1942-2466.
\newblock \doi{10.1029/2019ms002015}.
\newblock URL \url{https://dx.doi.org/10.1029/2019MS002015}.

\bibitem[Eyring et~al.(2016)Eyring, Bony, Meehl, Senior, Stevens, Stouffer, and Taylor]{eyring2016overview}
Eyring, V., Bony, S., Meehl, G.~A., Senior, C.~A., Stevens, B., Stouffer, R.~J., and Taylor, K.~E.
\newblock Overview of the coupled model intercomparison project phase 6 (cmip6) experimental design and organization.
\newblock \emph{Geoscientific Model Development}, 9\penalty0 (5):\penalty0 1937--1958, 2016.

\bibitem[Fay \& McKinley(2014)Fay and McKinley]{fay2014global}
Fay, A. and McKinley, G.
\newblock Global open-ocean biomes: mean and temporal variability.
\newblock \emph{Earth System Science Data}, 6\penalty0 (2):\penalty0 273--284, 2014.

\bibitem[Frölicher et~al.(2016)Frölicher, Rodgers, Stock, and Cheung]{RN164}
Frölicher, T.~L., Rodgers, K.~B., Stock, C.~A., and Cheung, W. W.~L.
\newblock Sources of uncertainties in 21st century projections of potential ocean ecosystem stressors.
\newblock \emph{Global Biogeochemical Cycles}, 30\penalty0 (8):\penalty0 1224--1243, 2016.
\newblock ISSN 0886-6236.

\bibitem[Garcia et~al.(2010)Garcia, Locarnini, Boyer, Antonov, Baranova, Zweng, and Johnson]{garcia2010world}
Garcia, H.~E., Locarnini, R.~A., Boyer, T.~P., Antonov, J.~I., Baranova, O.~K., Zweng, M.~M., and Johnson, D.~R.
\newblock \emph{World Ocean Atlas 2009, Volume 3: Dissolved Oxygen, Apparent Oxygen Utilization, and Oxygen Saturation}.
\newblock Number~70 in NOAA Atlas NESDIS. U.S. Government Printing Office, Washington, D.C., 2010.

\bibitem[Giglio et~al.(2018)Giglio, Lyubchich, and Mazloff]{Giglio2018EstimatingOI}
Giglio, D., Lyubchich, V., and Mazloff, M.~R.
\newblock Estimating oxygen in the southern ocean using argo temperature and salinity.
\newblock \emph{Journal of Geophysical Research: Oceans}, 2018.

\bibitem[Gong et~al.(2021)Gong, Li, and Zhou]{RN163}
Gong, H., Li, C., and Zhou, Y.
\newblock Emerging global ocean deoxygenation across the 21st century.
\newblock \emph{Geophysical Research Letters}, 48\penalty0 (23), 2021.
\newblock ISSN 0094-8276.

\bibitem[Ha et~al.(2017)Ha, Dai, and Le]{DBLP:conf/iclr/HaDL17}
Ha, D., Dai, A.~M., and Le, Q.~V.
\newblock Hypernetworks.
\newblock In \emph{5th International Conference on Learning Representations, {ICLR} 2017, Toulon, France, April 24-26, 2017, Conference Track Proceedings}, 2017.

\bibitem[Hamilton et~al.(2017)Hamilton, Ying, and Leskovec]{DBLP:conf/nips/HamiltonYL17}
Hamilton, W.~L., Ying, Z., and Leskovec, J.
\newblock Inductive representation learning on large graphs.
\newblock In \emph{Advances in Neural Information Processing Systems 30: Annual Conference on Neural Information Processing Systems 2017, December 4-9, 2017, Long Beach, CA, {USA}}, pp.\  1024--1034, 2017.

\bibitem[He et~al.(2019)He, Kw, Cr, I, Tp, Mm, Mm, Av, Ok, D, and Jr]{WOA18DO}
He, G., Kw, W., Cr, P., I, S., Tp, B., Mm, L., Mm, Z., Av, M., Ok, B., D, S., and Jr, R.
\newblock World ocean atlas 2018, volume 3: Dissolved oxygen, apparent oxygen utilization, and dissolved oxygen saturation.
\newblock Report, 2019.
\newblock URL \url{https://archimer.ifremer.fr/doc/00651/76337/}.

\bibitem[Huang et~al.(2023)Huang, Shao, Chen, Qi, Wu, Zhang, He, and Du]{Huang2023ReconstructionOD}
Huang, S., Shao, J., Chen, Y., Qi, J., Wu, S., Zhang, F., He, X., and Du, Z.
\newblock Reconstruction of dissolved oxygen in the indian ocean from 1980 to 2019 based on machine learning techniques.
\newblock \emph{Frontiers in Marine Science}, 2023.

\bibitem[Hussain et~al.(2021)Hussain, Durand, Coffin, Valdés, and Poirier]{hussain2021data}
Hussain, M.~M., Durand, G., Coffin, M., Valdés, J.~J., and Poirier, L.
\newblock Data driven study of estuary hypoxia.
\newblock In \emph{NeurIPS 2021 Workshop on Tackling Climate Change with Machine Learning}, 2021.
\newblock URL \url{https://www.climatechange.ai/papers/neurips2021/44}.

\bibitem[Ishizu \& Richards(2013)Ishizu and Richards]{ishizu2013relationship}
Ishizu, M. and Richards, K.~J.
\newblock Relationship between oxygen, nitrate, and phosphate in the world ocean based on potential temperature.
\newblock \emph{Journal of Geophysical Research: Oceans}, 118\penalty0 (7):\penalty0 3586--3594, 2013.

\bibitem[Jin et~al.(2023)Jin, Koh, Wen, Zambon, Alippi, Webb, King, and Pan]{DBLP:journals/corr/abs-2307-03759}
Jin, M., Koh, H.~Y., Wen, Q., Zambon, D., Alippi, C., Webb, G.~I., King, I., and Pan, S.
\newblock A survey on graph neural networks for time series: Forecasting, classification, imputation, and anomaly detection.
\newblock \emph{CoRR}, abs/2307.03759, 2023.

\bibitem[Kipf \& Welling(2017)Kipf and Welling]{GCN}
Kipf, T.~N. and Welling, M.
\newblock Semi-supervised classification with graph convolutional networks.
\newblock In \emph{5th International Conference on Learning Representations, {ICLR} 2017, Toulon, France, April 24-26, 2017, Conference Track Proceedings}. OpenReview.net, 2017.
\newblock URL \url{https://openreview.net/forum?id=SJU4ayYgl}.

\bibitem[Koelling et~al.(2023)Koelling, Atamanchuk, Wallace, and Karstensen]{Koelling2023DecadalVO}
Koelling, J., Atamanchuk, D., Wallace, D. W.~R., and Karstensen, J.
\newblock Decadal variability of oxygen uptake, export, and storage in the labrador sea from observations and cmip6 models.
\newblock \emph{Frontiers in Marine Science}, 2023.

\bibitem[Lam et~al.(2023)Lam, Sanchez-Gonzalez, Willson, Wirnsberger, Fortunato, Alet, Ravuri, Ewalds, Eaton-Rosen, Hu, Merose, Hoyer, Holland, Vinyals, Stott, Pritzel, Mohamed, and Battaglia]{doi:10.1126/science.adi2336}
Lam, R., Sanchez-Gonzalez, A., Willson, M., Wirnsberger, P., Fortunato, M., Alet, F., Ravuri, S., Ewalds, T., Eaton-Rosen, Z., Hu, W., Merose, A., Hoyer, S., Holland, G., Vinyals, O., Stott, J., Pritzel, A., Mohamed, S., and Battaglia, P.
\newblock Learning skillful medium-range global weather forecasting.
\newblock \emph{Science}, pp.\  eadi2336, 2023.

\bibitem[Li et~al.(2023)Li, Xia, Xu, and Huang]{DBLP:conf/nips/LiXX023}
Li, Z., Xia, L., Xu, Y., and Huang, C.
\newblock {GPT-ST:} generative pre-training of spatio-temporal graph neural networks.
\newblock In \emph{Advances in Neural Information Processing Systems 36: Annual Conference on Neural Information Processing Systems 2023, NeurIPS 2023, New Orleans, LA, USA, December 10 - 16, 2023}, 2023.

\bibitem[Long et~al.(2021)Long, Moore, Lindsay, Levy, Doney, Luo, Krumhardt, Letscher, Grover, and Sylvester]{Long2021SimulationsWT}
Long, M.~C., Moore, J., Lindsay, K., Levy, M.~N., Doney, S.~C., Luo, J.~Y., Krumhardt, K.~M., Letscher, R.~T., Grover, M.~A., and Sylvester, Z.~T.
\newblock Simulations with the marine biogeochemistry library (marbl).
\newblock \emph{Journal of Advances in Modeling Earth Systems}, 13, 2021.

\bibitem[Lu et~al.(2020)Lu, Gan, Jin, Fu, and Zhang]{DBLP:conf/cikm/LuGJFZ20}
Lu, B., Gan, X., Jin, H., Fu, L., and Zhang, H.
\newblock Spatiotemporal adaptive gated graph convolution network for urban traffic flow forecasting.
\newblock In \emph{{CIKM} '20: The 29th {ACM} International Conference on Information and Knowledge Management, Virtual Event, Ireland, October 19-23, 2020}, pp.\  1025--1034. {ACM}, 2020.

\bibitem[Lu et~al.(2022)Lu, Gan, Zhang, Yao, Fu, and Wang]{DBLP:conf/kdd/0005G0YFW22}
Lu, B., Gan, X., Zhang, W., Yao, H., Fu, L., and Wang, X.
\newblock Spatio-temporal graph few-shot learning with cross-city knowledge transfer.
\newblock In \emph{{KDD} '22: The 28th {ACM} {SIGKDD} Conference on Knowledge Discovery and Data Mining, Washington, DC, USA, August 14 - 18, 2022}, pp.\  1162--1172. {ACM}, 2022.

\bibitem[Lu et~al.(2023)Lu, Wu, Yang, Sun, Liu, Gan, Liang, Fu, Wang, and Zhou]{10.1145/3543873.3587305}
Lu, B., Wu, L., Yang, L., Sun, C., Liu, W., Gan, X., Liang, S., Fu, L., Wang, X., and Zhou, C.
\newblock Dataexpo: A one-stop dataset service for open science research.
\newblock In \emph{Companion Proceedings of the ACM Web Conference 2023}, WWW '23 Companion, pp.\  32–36, 2023.

\bibitem[Miao et~al.(2023)Miao, Wu, Chen, Gao, and Yin]{DBLP:journals/tkde/MiaoWCGY23}
Miao, X., Wu, Y., Chen, L., Gao, Y., and Yin, J.
\newblock An experimental survey of missing data imputation algorithms.
\newblock \emph{{IEEE} Trans. Knowl. Data Eng.}, 35\penalty0 (7):\penalty0 6630--6650, 2023.

\bibitem[Muller et~al.(2019)Muller, Lange, Wang, Piorkowski, Tsay, Liao, Dugan, and Erickson]{10.1145/3290605.3300356}
Muller, M., Lange, I., Wang, D., Piorkowski, D., Tsay, J., Liao, Q.~V., Dugan, C., and Erickson, T.
\newblock How data science workers work with data: Discovery, capture, curation, design, creation.
\newblock In \emph{Proceedings of the 2019 CHI Conference on Human Factors in Computing Systems}, CHI '19, pp.\  1–15, New York, NY, USA, 2019.

\bibitem[Nguyen et~al.(2023)Nguyen, Brandstetter, Kapoor, Gupta, and Grover]{Nguyen2023ClimaXAF}
Nguyen, T., Brandstetter, J., Kapoor, A., Gupta, J.~K., and Grover, A.
\newblock Climax: A foundation model for weather and climate.
\newblock In \emph{International Conference on Machine Learning}, 2023.

\bibitem[Ning et~al.(2023)Ning, Vetrova, and Bryan]{ning2023graphbased}
Ning, D., Vetrova, V., and Bryan, K.
\newblock Graph-based deep learning for sea surface temperature forecasts.
\newblock In \emph{ICLR 2023 Workshop on Tackling Climate Change with Machine Learning}, 2023.
\newblock URL \url{https://www.climatechange.ai/papers/iclr2023/39}.

\bibitem[{NOAA National Centers for Environmental Information}(2022)]{etopo2022}
{NOAA National Centers for Environmental Information}.
\newblock Etopo 2022 60 arc-second global relief model, 2022.
\newblock URL \url{https://www.ncei.noaa.gov/products/etopo-global-relief-model}.

\bibitem[Pal et~al.(2021)Pal, Ma, Zhang, and Coates]{DBLP:conf/icml/PalMZC21}
Pal, S., Ma, L., Zhang, Y., and Coates, M.
\newblock {RNN} with particle flow for probabilistic spatio-temporal forecasting.
\newblock In \emph{Proceedings of the 38th International Conference on Machine Learning, {ICML} 2021, 18-24 July 2021, Virtual Event}, volume 139 of \emph{Proceedings of Machine Learning Research}, pp.\  8336--8348. {PMLR}, 2021.

\bibitem[Pant et~al.(2023)Pant, Toshniwal, and Gurjar]{10.1007/978-3-031-24378-3_7}
Pant, N., Toshniwal, D., and Gurjar, B.~R.
\newblock Application of attention mechanism combined with long short-term memory for forecasting dissolved oxygen in ganga river.
\newblock In \emph{Advanced Analytics and Learning on Temporal Data: 7th ECML PKDD Workshop, AALTD 2022, Grenoble, France, September 19–23, 2022, Revised Selected Papers}, pp.\  105–116, Berlin, Heidelberg, 2023. Springer-Verlag.
\newblock ISBN 978-3-031-24377-6.

\bibitem[Podina et~al.(2023)Podina, Eastman, and Kohandel]{DBLP:conf/icml/PodinaEK23}
Podina, L., Eastman, B., and Kohandel, M.
\newblock Universal physics-informed neural networks: Symbolic differential operator discovery with sparse data.
\newblock In \emph{International Conference on Machine Learning, {ICML} 2023, 23-29 July 2023, Honolulu, Hawaii, {USA}}, volume 202 of \emph{Proceedings of Machine Learning Research}, pp.\  27948--27956. {PMLR}, 2023.

\bibitem[Raissi et~al.(2019)Raissi, Perdikaris, and Karniadakis]{DBLP:journals/jcphy/RaissiPK19}
Raissi, M., Perdikaris, P., and Karniadakis, G.~E.
\newblock Physics-informed neural networks: {A} deep learning framework for solving forward and inverse problems involving nonlinear partial differential equations.
\newblock \emph{J. Comput. Phys.}, 378:\penalty0 686--707, 2019.

\bibitem[Redfield et~al.(1963)Redfield, Ketchum, Richards, et~al.]{redfield1963influence}
Redfield, A., Ketchum, B., Richards, F., et~al.
\newblock The influence of organisms on the composition of seawater.
\newblock \emph{The sea}, 2:\penalty0 26--77, 1963.

\bibitem[Reichstein et~al.(2019)Reichstein, Camps-Valls, Stevens, Jung, Denzler, Carvalhais, and Prabhat]{Reichstein2019DeepLA}
Reichstein, M., Camps-Valls, G., Stevens, B., Jung, M., Denzler, J., Carvalhais, N., and Prabhat.
\newblock Deep learning and process understanding for data-driven earth system science.
\newblock \emph{Nature}, 566:\penalty0 195 -- 204, 2019.

\bibitem[Reygondeau et~al.(2020)Reygondeau, Cheung, Wabnitz, Lam, Fr{\"o}licher, and Maury]{Reygondeau2020ClimateCE}
Reygondeau, G., Cheung, W. W.~L., Wabnitz, C. C.~C., Lam, V. W.~Y., Fr{\"o}licher, T.~L., and Maury, O.
\newblock Climate change-induced emergence of novel biogeochemical provinces.
\newblock In \emph{Frontiers in Marine Science}, 2020.

\bibitem[Schmidtko et~al.(2017)Schmidtko, Stramma, and Visbeck]{Schmidtko2017DeclineIG}
Schmidtko, S., Stramma, L., and Visbeck, M.
\newblock Decline in global oceanic oxygen content during the past five decades.
\newblock \emph{Nature}, 542:\penalty0 335--339, 2017.

\bibitem[Sonnewald et~al.(2020)Sonnewald, Dutkiewicz, Hill, and Forget]{Sonnewald2020ElucidatingEC}
Sonnewald, M., Dutkiewicz, S., Hill, C.~N., and Forget, G.
\newblock Elucidating ecological complexity: Unsupervised learning determines global marine eco-provinces.
\newblock \emph{Science Advances}, 6, 2020.

\bibitem[Stramma et~al.(2008)Stramma, Johnson, Sprintall, and Mohrholz]{doi:10.1126/science.1153847}
Stramma, L., Johnson, G.~C., Sprintall, J., and Mohrholz, V.
\newblock Expanding oxygen-minimum zones in the tropical oceans.
\newblock \emph{Science}, 320\penalty0 (5876):\penalty0 655--658, 2008.
\newblock \doi{10.1126/science.1153847}.

\bibitem[Sun et~al.(2023)Sun, Cucuzzella, Brus, Narayanan, Nadiga, Van~Roekel, Hückelheim, and Madireddy]{sun2023surrogate}
Sun, Y., Cucuzzella, E., Brus, S., Narayanan, S. H.~K., Nadiga, B., Van~Roekel, L., Hückelheim, J., and Madireddy, S.
\newblock Surrogate neural networks to estimate parametric sensitivity of ocean models.
\newblock In \emph{NeurIPS 2023 Workshop on Tackling Climate Change with Machine Learning}, 2023.
\newblock URL \url{https://www.climatechange.ai/papers/neurips2023/65}.

\bibitem[Tobler(1970)]{tobler1970computer}
Tobler, W.~R.
\newblock A computer movie simulating urban growth in the detroit region.
\newblock \emph{Economic geography}, 46:\penalty0 234--240, 1970.

\bibitem[Velickovic et~al.(2018)Velickovic, Cucurull, Casanova, Romero, Li{\`{o}}, and Bengio]{DBLP:conf/iclr/VelickovicCCRLB18}
Velickovic, P., Cucurull, G., Casanova, A., Romero, A., Li{\`{o}}, P., and Bengio, Y.
\newblock Graph attention networks.
\newblock In \emph{6th International Conference on Learning Representations, {ICLR} 2018, Vancouver, BC, Canada, April 30 - May 3, 2018, Conference Track Proceedings}. OpenReview.net, 2018.
\newblock URL \url{https://openreview.net/forum?id=rJXMpikCZ}.

\bibitem[Wang et~al.(2023)Wang, Yan, Qiu, Zhu, Guan, Margenot, and Tong]{DBLP:conf/kdd/WangYQZGMT23}
Wang, D., Yan, Y., Qiu, R., Zhu, Y., Guan, K., Margenot, A., and Tong, H.
\newblock Networked time series imputation via position-aware graph enhanced variational autoencoders.
\newblock In \emph{Proceedings of the 29th {ACM} {SIGKDD} Conference on Knowledge Discovery and Data Mining, {KDD} 2023, Long Beach, CA, USA, August 6-10, 2023}, pp.\  2256--2268. {ACM}, 2023.

\bibitem[Yik et~al.(2023)Yik, Sonnewald, Clare, and Lguensat]{yik2023southern}
Yik, W.~J., Sonnewald, M., Clare, M., and Lguensat, R.
\newblock Southern ocean dynamics under climate change: New knowledge through physics-guided machine learning.
\newblock In \emph{NeurIPS 2023 Workshop on Tackling Climate Change with Machine Learning}, 2023.
\newblock URL \url{https://www.climatechange.ai/papers/neurips2023/37}.

\bibitem[Yoon et~al.(2018)Yoon, Jordon, and van~der Schaar]{DBLP:conf/icml/YoonJS18}
Yoon, J., Jordon, J., and van~der Schaar, M.
\newblock {GAIN:} missing data imputation using generative adversarial nets.
\newblock In \emph{Proceedings of the 35th International Conference on Machine Learning, {ICML} 2018, Stockholmsm{\"{a}}ssan, Stockholm, Sweden, July 10-15, 2018}, volume~80 of \emph{Proceedings of Machine Learning Research}, pp.\  5675--5684. {PMLR}, 2018.

\bibitem[Zeng et~al.(2023)Zeng, Kothari, Bryngelson, and Sch{\"{a}}fer]{DBLP:conf/iclr/ZengKBS23}
Zeng, Q., Kothari, Y., Bryngelson, S.~H., and Sch{\"{a}}fer, F.
\newblock Competitive physics informed networks.
\newblock In \emph{The Eleventh International Conference on Learning Representations, {ICLR} 2023, Kigali, Rwanda, May 1-5, 2023}, 2023.

\bibitem[Zhou et~al.(2022)Zhou, Gong, and Zhou]{Zhou2022ResponsesOH}
Zhou, Y., Gong, H., and Zhou, F.
\newblock Responses of horizontally expanding oceanic oxygen minimum zones to climate change based on observations.
\newblock \emph{Geophysical Research Letters}, 49, 2022.

\end{thebibliography}
\bibliographystyle{icml2023}

\newpage
\appendix
\onecolumn

\section{Notations}
\label{appendix:notation}

In Table \ref{tab:symbol}, we list the main symbols and notations used throughout the paper. 

\begin{table}[h]
\centering
\caption{Symbols and Notations.}
\label{tab:symbol}
\begin{tabular}{c|l} 
\toprule
Symbol & Definition \\
\midrule
$\mathbf{X}$ & Data matrix of dissolved oxygen\\
$\hat{\mathbf{X}}$ & Estimated Data matrix of dissolved oxygen\\
$\textbf{X}^{(obs)}$ & Observed data matrix\\
$\textbf{X}^{(imp)}$ & Imputed data matrix\\
$\bm{\Omega}$ & Binary mask representing observed entries\\
$\mathbf{F}^{\text{Geo}}$ & Matrix of geographical factor\\
$\mathbf{F}^{\text{Env}}$ & Matrix of environmental factor\\
$\bm{\mathcal{E}}$ & Edge feature matrix\\
NA & Indicator of not available data observation\\
\midrule
$X_t$ & Dissolved oxygen (DO) observation at time $t$\\
$X_{t-T:t+T}$ & DO observation from time $t-T$ to $t+T$\\
$F_t^{\text{Geo}}$ & Geographical coordinates at time $t$\\
$F_t^{\text{Env}}$ & Environmental factors at time $t$\\
${A}_t$ & Adjacency matrix at time $t$\\
$\mathcal{V}_t$ & Node set at time $t$\\
$\mathcal{E}_t$ & Edge feature matrix at time $t$\\
$Z_t$ & Latent embedding matrix of feature extractor at time $t$\\
$f(\theta)$ & Feature extractor parameterized by $\theta$\\
$A_\pi$ & HyperNetwork for parameter $\alpha$ and $\beta$\\
\midrule
L, G, D, T & Number of longitude, latitude, depth and time interval\\
i, j, d, t & Indices of graph nodes\\
$\mathcal{Z}^\alpha_{m}$ & Rotation matrix in hypernetwork\\
$\mathcal{Z}^\beta_{m}$ & Bias vector in hypernetwork\\
$v_{i,j,d,t}$ & Node with spatial coordinate $(i,j,t)$ at time $t$\\
$v_{m}$ & The $m$-th node for simplification\\
$e_{m,n}$ & The edge between node $v_{m}$ and $v_{n}$\\
$\gamma_{m,n}$ & The edge weight of the message-passing between node $v_{m}$ and $v_{n}$\\
$\xi_m$ & The context information of node $v_m$ \\
$\mathcal{N}_{m}$ & Neighbors of node $v_{m}$\\
$\tilde{\mathcal{N}}_{m}$ & Neighbors of node $v_{m}$ with self-loop\\
$\odot$ & Element-wise product\\
$\sigma(\cdot)$ & Variance\\
$\|$ & Feature concatenation\\
\bottomrule
\end{tabular}
\end{table}

\section{Background of Ocean Deoxygenation}

\textbf{Drivers and Impacts.} Ocean deoxygenation is a complex environmental issue characterized by the significant reduction in dissolved oxygen levels in the ocean. Several interconnected factors, especially the climate change and anthropogenic activities, contribute to ocean deoxygenation. Rising temperatures impact oxygen circulation and diminish the capacity of seawater to hold dissolved oxygen, resulting in decreased oceanic oxygen levels. Excessive nutrient input from human activities, particularly from agriculture and industrial processes, lead to the proliferation of algae, creating algal blooms. When these blooms eventually decompose, they consume large amounts of oxygen, creating localized areas with reduced oxygen levels. In recent years, with the intensification of global warming and human influence, ocean deoxygenation has presented an accelerating trend. The consequences of ocean deoxygenation are far-reaching. Reduced oxygen levels can harm marine biodiversity, leading to disruptions in food webs and ecosystems. Fisheries, which rely on oxygen-rich environments to support commercially important species, may experience declines. Hence, there is an urgent need to understand the mechanisms driving deoxygenation, assess its ecological consequences, and develop effective strategies for sustainable development.

\textbf{El Niño-Southern Oscillation (ENSO).} Anomalous climate events influence the distribution and rates of ocean deoxygenation. Taking the El Niño-Southern Oscillation (ENSO) as an example, this climatic phenomenon greatly disturbs oceanic dissolved oxygen levels. During El Niño years, the warming of the central and eastern equatorial Pacific reduces the solubility of oxygen, leading to a decline in dissolved oxygen concentrations in affected areas. Understanding the intricate relationship between ENSO and oceanic dissolved oxygen is crucial for comprehending the broader implications of climate-related variations on marine ecosystems.

\textbf{Oxygen Minimum Zones (OMZs).} Oxygen Minimum Zones (OMZs), standing for low oxygen zones, plays a pivotal role in characterizing the issue of oceanic oxygen deficiency. OMZs are emerging in various regions, posing challenges for marine organisms adapted to higher oxygen concentrations. In various studies, OMZs are bounded by different thresholds of dissolved oxygen levels. In our work, we select a 30 $\mu$mol/kg threshold to define OMZs, referred to as OMZ30. To be specific, areas with dissolved oxygen concentrations below 30 $\mu$mol/kg at any depth are labeled as OMZ30. In Figure \ref{figure1}, OMZ30 are highlighted by yellow lines and our work indicates the significant expansion of OMZ30 over the past century. This observation signifies a clear trend of ocean deoxygenation, emphasizing the urgent need for in-depth investigations into deoxygenation rates and driving factors.

\section{Data Sources}

\subsection{Observation Data}
\label{appendix:observ_data}

In our work, we aggregate comprehensive observation datasets of global ocean from 5 publicly available database. The circulation of dissolved oxygen in seawater is a complex biological, physical, and chemical effect. Therefore, besides global dissolved oxygen (DO) measurements, we also combine more than 100-year observations of seawater temperature, salinity, biogeochemical major elements (P, N, Si) and chlorophyll. In Table \ref{tab:data-sources}, we demonstrate the detailed information of different data sources. Moreover, in our analysis, we also use bedrock elevation data~\cite{etopo2022} to represent the changes in seabed depth in different regions. 

\begin{table}[h]
\centering
\caption{Detailed Information of Data sources for global ocean observations.}
\label{tab:data-sources}
\renewcommand{\arraystretch}{1.1}
\resizebox{\textwidth}{!}{%
\begin{tabular}{@{}llllll@{}}
\toprule
Database &
  Time &
  Institution &
  Source &
  Access Date &
  Variables \\ \midrule
\begin{tabular}[c]{@{}l@{}}World Ocean Database\\ (WOD 2018)\end{tabular} &
  1900-2023 &
  \begin{tabular}[c]{@{}l@{}}National Centers for\\ Environmental Information\end{tabular} &
  \url{https://www.ncei.noaa.gov/} &
  2023-05 &
  \begin{tabular}[c]{@{}l@{}}temperature, salinity, dissolved\\ oxygen, biogeochemical major\\ elements (P, N, Si) and chlorophyll.\end{tabular} \\
\begin{tabular}[c]{@{}l@{}}CLIVAR and Carbon \\ Hydrographic Database\\ (CCHDO)\end{tabular} &
  1922-2023 &
  \begin{tabular}[c]{@{}l@{}}CLIVAR and Carbon\\ Hydrographic Data Office\end{tabular} &
  \url{https://cchdo.ucsd.edu/} &
  2023-05 &
  \begin{tabular}[c]{@{}l@{}}temperature, salinity, dissolved\\ oxygen, and  biogeochemical major\\ elements (P, N, Si).\end{tabular} \\
Argo &
  2001-2023 &
  \begin{tabular}[c]{@{}l@{}}Argo Global Data\\ Assemby Center\end{tabular} &
  \url{https://argo.ucsd.edu/} &
  2023-05 &
  \begin{tabular}[c]{@{}l@{}}temperature, salinity, dissolved\\ oxygen, biogeochemical major\\ elements (N) and chlorophyll.\end{tabular} \\
\begin{tabular}[c]{@{}l@{}}Global Ocean Data Analysis\\ Project version2.2022\\ (GLODAPV2\_2022)\end{tabular} &
  1972-2021 &
  \begin{tabular}[c]{@{}l@{}}NOAA’s National Centers\\ for Environmental\\ Information (NCEI)\end{tabular} &
  \url{https://glodap.info/} &
  2023-05 &
  \begin{tabular}[c]{@{}l@{}}temperature, salinity, dissolved\\ oxygen, biogeochemical major\\ elements (P, N, Si) and chlorophyll.\end{tabular} \\
Geotraces IDP &
  2007-2018 &
  \begin{tabular}[c]{@{}l@{}}GEOTRACES International\\ Data Assembly Centre\\ (GDAC)\end{tabular} &
  \url{https://www.geotraces.org} &
  2023-10 &
  \begin{tabular}[c]{@{}l@{}}temperature, salinity, dissolved\\ oxygen, biogeochemical major\\ elements (P, N, Si) and chlorophyll.\end{tabular} \\ \bottomrule
\end{tabular}%
}
\end{table}

In addition, to ensure the availability and accuracy of the analysis data from various sources, we have set unified quality control standards. Quality control in oceanic research is a multifaceted process that involves careful attention to instrument calibration, data validation, metadata documentation, and various other checks to ensure the reliability and accuracy of collected data.
"FLAG" is a marker associated with a specific data point, which helps researchers, analysts, or automated systems identify potential issues, outliers, or anomalies in the dataset.
In our paper, we unify the flag system of multiple databases and form the following quality control standards as shown in Table \ref{tab:flag}.

\begin{table}[]
\centering
\caption{Flag system of multiple databases.}
\label{tab:flag}
\renewcommand{\arraystretch}{1.1}
\begin{tabular}{lccccc}
\hline
\multicolumn{1}{c}{FLAG} & WOD             & CCHDO   & Argo & GLODAPV2\_2022 & IDP 2021        \\ \hline
0: good quality          & 0               & 2       & 1    & 2              & 1               \\
1: unknown quality       & N/A             & 0,1,5,8 & N/A  & N/A            & 0,5             \\
2: questionable quality  & 1               & 3,6,7   & 2,3  & 0              & 2,3,6,7,8,A,B,Q \\
3: bad quality           & 2,3,4,5,6,7,8,9 & 4       & 4    & N/A            & 4               \\
4: not sampled           & N/A             & 9       & 0,5  & 9              & 9               \\ \hline
\end{tabular}%
\end{table}

\begin{figure}
    \centering
    \includegraphics[width=0.92\linewidth]{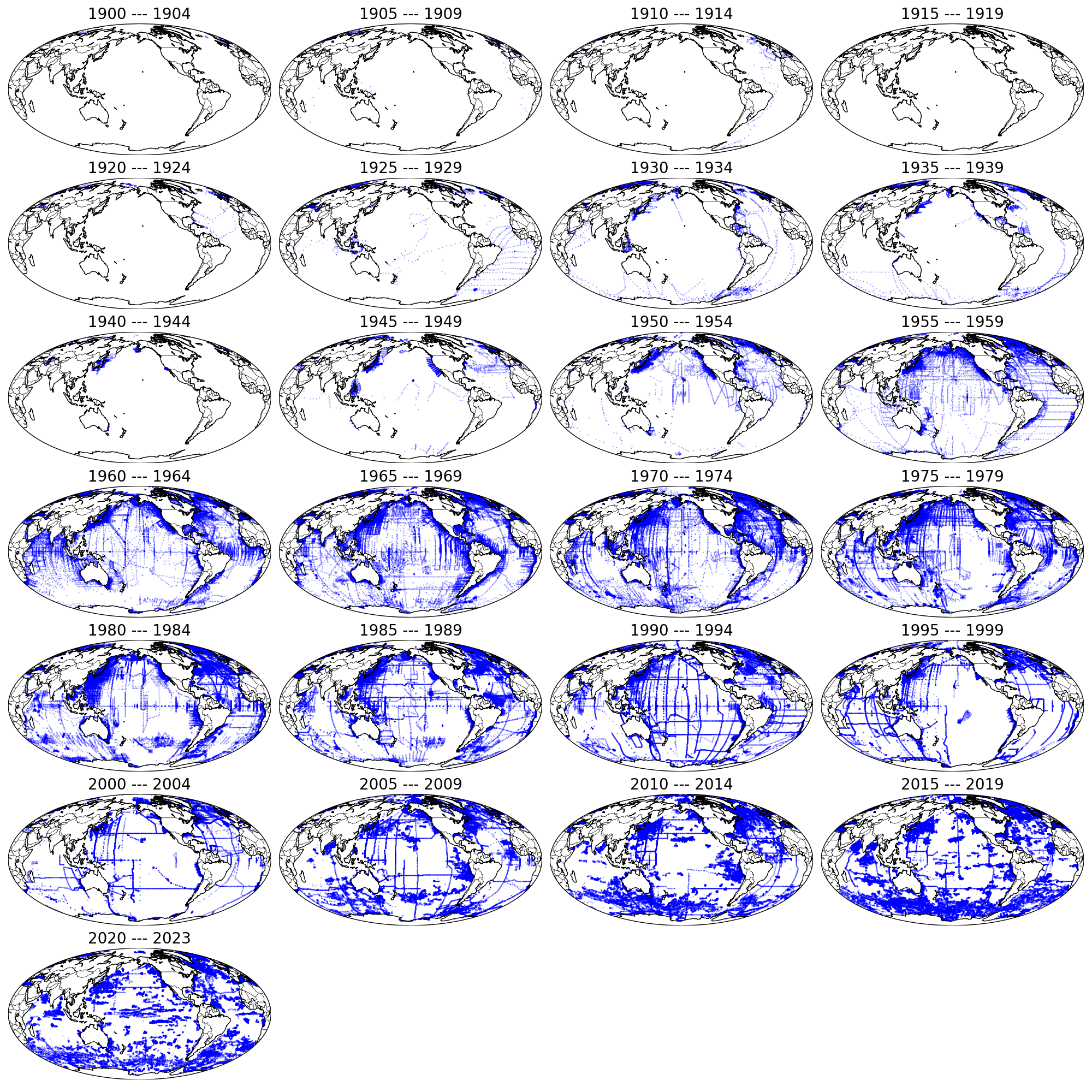}
    \caption{The spatio-temporal distribution of dissolved oxygen observation data collected from multiple databases.}
    \label{fig:data-collection}
\end{figure}

\subsection{Simulation Data}
\label{appendix:cmip6}
To demonstrate the effectiveness of our sparse observations reconstruction, we acquire three sets of experimental data from the Coupled Model Inter-comparison Project Phase 6 (CMIP6). 
The primary goal of CMIP is to provide a framework for comparing and evaluating climate models to better understand and predict future climate changes. CMIP6 is the latest phase in a series of such projects and was designed to support the assessment reports of the Intergovernmental Panel on Climate Change (IPCC). Under CMIP6, participating modeling centers simulate the Earth's climate system using a standardized set of protocols and scenarios.  CMIP6 plays a crucial role in advancing our understanding of climate processes, improving model performance, and providing policymakers with more reliable information about potential future climate scenarios.

In our work, we select experiments involve historical simulations of ocean dissolved oxygen distribution, accessed from \url{https://esgf-node.llnl.gov/search/cmip6/}. We conduct performance evaluations for three sets of simulation data respectively and carry out comparative analyses with our reconstruction results. All three datasets include four-dimensional spatio-temporal information, longitude, latitude, depth, and time. Besides, they provide simulation results for dissolved oxygen concentration during historical periods. Particularly, these datasets are characterized by a spatial resolution of $1^\circ \times 1^\circ$, representing annual averages. And the depth level bounds are consistent with our data grids, facilitating direct comparisons. The main information for these three experiments is provided in Table \ref{tab:simul-data-info}, with detailed descriptions as follows:
\begin{itemize} 
    \item \textbf{CESM2 omip1}: The Community Earth System Model (CESM) is a climate/Earth system coupled model used to simulate past, present, and future climates. The ocean component of CESM2 undergoes various physical model and numerical computation improvements, while utilizing the Marine Biogeochemistry Library (MARBL) to represent ocean biogeochemistry. The experiment, omip1, is driven by the CORE-II (Coordinated Ocean - ice Reference Experiments) atmospheric data, and initialized with physical and biogeochemical ocean observations to conduct ocean dissolved oxygen simulations.
    \item \textbf{CESM2 omip2}: Experiment omip2 shares the same model configuration as omip1 but is forced by the JRA-55 atmospheric data with higher spatial resolution than CORE-II.
    \item \textbf{GFDL-ESM4 historical}: The Earth System Model4 (ESM4) is the fourth-generation chemistry-carbon-climate coupled climate model developed by the Geophysical Fluid Dynamics Laboratory (GFDL). Compared to previous versions, this model more comprehensively represents chemical cycling and ecosystems. Particularly, the model incorporates interactions between ocean ecology and biogeochemistry. The historical experiment utilizes rich climate observational data from 1850 to the present, imposing environmental change conditions consistent with observations to obtain historical simulations of ocean dissolved oxygen.
\end{itemize}

\begin{table}[h]
\centering
\caption{Detailed Information of Simulation Data from CMIP6.}
\label{tab:simul-data-info}
\renewcommand{\arraystretch}{1.1}
\resizebox{\textwidth}{!}{%
\begin{tabular}{@{}llllll@{}}
\toprule
Model &
  Experiment &
  Institution &
  Access Date \\ \midrule
\begin{tabular}[c]{@{}l@{}}Community Earth System Model (CESM2)\\ \cite{CESM2}\end{tabular} &
  omip1&
  \begin{tabular}[c]{@{}l@{}}National Center for Atmospheric Research\\(NCAR)\end{tabular} &
  2023-11\\
\begin{tabular}[c]{@{}l@{}}Community Earth System Model (CESM2) \\ \cite{CESM2}\end{tabular} &
  omip2 &
  \begin{tabular}[c]{@{}l@{}}National Center for Atmospheric Research\\(NCAR)\end{tabular} &
  2023-11\\
\begin{tabular}[c]{@{}l@{}}Geophysical Fluid Dynamics Laboratory\\ - Earth System Model(GFDL-ESM4)\\~\cite{GFDL-ESM4}\end{tabular} &
  historical &
  \begin{tabular}[c]{@{}l@{}}NOAA's Geophysical Fluid Dynamics Laboratory\\(GFDL)\end{tabular} &
  2023-11\\ \bottomrule
\end{tabular}%
}
\end{table}

\subsection{Variables}

In this subsection, we conclude the input variables in our experiments as shown in Table \ref{tab:appendix-variable}. We provide detailed variable names, units, numerical ranges, and notes. With respect to the spatial coordinates, we convert the longitude and latitude into spherical coordinates to achieve the continuity of spatial geoemtric relationships and distance measurement.

\begin{table}[]
\centering
\caption{Description of variables in ocean deoxygenation reconstruction. }
\label{tab:appendix-variable}
\begin{threeparttable}
\resizebox{\textwidth}{!}{%
\begin{tabular}{@{}llrl@{}}
\toprule
Variable & Unit & Numeric Range & Notes \\ \midrule
Longitude & ° & -180-180 &  \\
Latitude & ° & -90-90 &  \\
Depth & m & 0-5500 &  \\
Time & Year & 1920-2023 &  \\
Pressure & dbar & 0-5500 & Calculated via International Equation of State of Seawater (TEOS-10) \\
Density & kg/$m^3$ & 1020-1040 & Calculated via International Equation of State of Seawater (TEOS-10) \\
Bathymetry & m & -8000-5500 & Refer to ETOPO Global Relief Model \\
Dissolved Oxygen & $\mu$mol/kg & 0-523 &  \\
Temperature & ℃ & -3-35 &  \\
Salinity & unitless & 0-44 &  \\
Nitrate & $\mu$mol/kg & 0-500 & WOD database does not differentiate between nitrate, nitrate+nitrite.\tnote{4} \\
Phosphate & $\mu$mol/kg & 0-5 & Except for Mediterranean, Black Sea, Baltic Sea \\
Silicate & $\mu$mol/kg & 0-250 & Except for Black Sea \\
Chlorophyll & $\mu$g/L & 0-50 &  \\
Chlorophyll A & $\mu$g/L & 0-50 &  \\ \bottomrule
\end{tabular}%
}
\begin{tablenotes}
\item [4] The content of nitrite in seawater is much lower than that of nitrate~\cite{WOA18DO,wod2018}.
\end{tablenotes}
\end{threeparttable}
\end{table}

\section{Related Work}
\label{appendix:related_work}

In this section, we conclude the related works to global ocean oxygenation reconstruction.

\paragraph{Ocean Deoxygenation.}
Global warming and excessive nutrient inputs caused by human activities have led to significant ocean deoxygenation in recent years. The loss of oxygen progressively diminishes the ocean's capacity to sustain high productivity and diverse biological communities, impacting both its economic and ecological functions~\cite{doi:10.1126/science.aam7240}. Detailed investigations into oxygen loss have been conducted in selected regions with abundant observational data dating back to the 1930s, such as equatorial Pacific, eastern tropical Atlantic~\cite{doi:10.1126/science.1153847}, and coastal northwest Atlantic~\cite{RN162}.

Furthermore, to better understand the oxygen cycling mechanism and assess the overall impact of human activities on the marine system since the 20th century, a comprehensive global analysis of ocean deoxygenation is essentially important. The main challenge lies in accurately characterizing the correlations between missing values and sparse observations to achieve global century-scale dissolved oxygen reconstruction. Existing research has made two types of positive attempts, but still has some limitations: (1) Many studies utilize numerical simulations based on climate models to probe the drivers and predict oxygen loss~\cite{RN163, RN165}. For example, Coupled Model Intercomparison Project Phase 6 (CMIP6)~\cite{eyring2016overview} conduct three experiments (CESM2-omip1, CESM2-omip2 and GFDL-ESM4-historical) on dissolved oxygen simulation.
Nevertheless, most simulations entirely rely on knowledge of the climate system and fail to leverage observations for correction, thereby showing inferior performance~\cite{RN164, RN166}. 
(2) To further utilize oceanic observations, there only exists a few studies attempt data reconstruction through fast marching algorithm~\cite{Schmidtko2017DeclineIG} and geostatistical regression~\cite{Zhou2022ResponsesOH}. Yet, these methods are unable to characterize complex spatio-temporal correlations and deoxygenation mechanism, thus limiting their imputation resolution and accuracy. 
In this paper, to the best of our knowledge, we are the first to propose a deep learning method to reconstruct global ocean deoxygenation over a century, considering the spatio-temporal hetegeneity in different regions (Section \ref{sec:4D-st-hypernetwork}) and chemical properties across dissovled oxygen and nutrients (Section \ref{sec:chem-grad-norm}).

\paragraph{Data-Driven Earth System.}
Existing superior earth system methods are mostly physics-informed numerical models, which relies on complicated physics process, sensitive initial condition and suitable forcing. Moreover, many such numerical models are computationally intensive and costly for fine-grained spatio-temporal resolution or long-range simulation. With the rapid growth of artificial intelligence, \emph{AI for Science}, or more specifically data-driven Earth system, has become a hot topic in both Computer Science and Earth Geoscience~\cite{Bergen2019MachineLF,Reichstein2019DeepLA,ning2023graphbased,yik2023southern,sun2023surrogate}. 
Given the growth of scientific data, data-driven deep learning models are attempting to learn complex and nonlinear correlations in the Earth system, while reducing computational and application costs.
Especially in the field of numerical weather prediction (NWP), deep learning techniques are particularly suitable for improving its prediction performance due to their large amount of data, e.g. recent state-of-the-art methods Pangu-Weather~\cite{bi2023accurate}, GraphCast~\cite{doi:10.1126/science.adi2336}. Pangu-Weather~\cite{bi2023accurate} designs a 3D Earth Specific Transformer architecture and for the first time outperforms state-of-the-art numerical weather prediction (NWP) methods. Similarly, GraphCast~\cite{doi:10.1126/science.adi2336} combines Graph Neural Networks to predicts five Earth-surface variables and six atmospheric variables accurately. 
On the other hand, ~\citeauthor{Nguyen2023ClimaXAF} propose a foundation model for weather and climate called ClimaX, which extends Transformer architecture for more general weather and climate tasks. Overall, the data-driven Earth system is still in its early stages, and more scenarios and technologies are worth further exploration.

\paragraph{Data Imputation for Spatio-temporal Data.}

Recently, spatio-temporal data mining has been widely studied in scenarios such as urban traffic flow prediction, electricity consumption prediction, taxi demand forecasting~\cite{DBLP:conf/cikm/LuGJFZ20,DBLP:conf/icml/PalMZC21,DBLP:conf/kdd/0005G0YFW22,DBLP:conf/nips/LiXX023}. 
Data missing is prevailing in spatio-temporal data analysis, due to various reasons such as uneven placement of data collector, equipment failure, signal interruptions, etc. 
Especially for ocean observation data, due to the high cost and difficulty of ocean data collection, ocean observation data is highly sparse. Recently, several latest survey papers on data imputation have emerged, providing comprehensive analysis from different perspectives such as systematic experimental comparison~\cite{DBLP:journals/tkde/MiaoWCGY23}, Internet of Things (IoT) applications~\cite{DBLP:journals/csur/AdhikariJZHRAK23}, and spatio-temporal graph neural network methods~\cite{DBLP:journals/corr/abs-2307-03759}.
Traditional data imputation methods are mainly based on statistical methods, such as mean completion, matrix factorization, spline interpolation, etc. With the rise of a series of deep learning methods, such as time series analysis, graph neural networks, etc., a series of superior methods~\cite{DBLP:conf/nips/CaoWLZLL18,DBLP:conf/icml/YoonJS18,DBLP:conf/iclr/CiniMA22,DBLP:conf/kdd/WangYQZGMT23} have been proposed to overcome the complex and nonlinear correlation between missing data and observed data. 

It should be pointed that classical data imputation works involve manually masking approximately 20\% -80\% of the data to create missing values, and then comparing the imputation results with complete data for performance validation. However, ocean observation data is highly sparse, for example, the global ocean deoxygenation observation data studied in this paper is only 3.735\%, so we cannot adopt the same experimental setup for deoxygenation reconstruction.
Meanwhile, the latest experimental survey on data imputation empirically verify that as the missing rate increases, the completion performance will show a significant decrease.

\begin{quote}
    \emph{"We can observe that, \underline{with the increase of missing rate}, the imputation accuracy (in terms of ARMSE and AMAE) \underline{descends consistently} for each algorithm. When more and more values are missing, the observed information becomes less, making imputation algorithms \underline{less effective}."}~\cite{DBLP:journals/tkde/MiaoWCGY23}
\end{quote}

\begin{quote}
    \emph{"The continuous missing gap from 3\% to 10\% of the data in a variable is termed as a high missing gap. The continuous missing gap \underline{higher than 10\%} of the data in a variable is a \underline{very high missing gap}. $\cdots$ It is to be noted that most of the existing research conducts experiments to impute missing data stating that simulated or real data consists of 50\% of the missing data. Such missing data are based on the missing rate or amount, lacking the information about the missing gap."}~\cite{DBLP:journals/csur/AdhikariJZHRAK23}
\end{quote}

Therefore, when reconstructing the concentration of DO at time $t$, we fully utilize the dissolved oxygen observation data from $t-1$ to $t-T$ and from $t+1$ to $t+T$ timesteps, as well as auxiliary variables at time $t$. Then, we evaluate the reconstruction performance using observations at time $t$.

\section{More Details about \textsc{OxyGenerator}}

\subsection{Feature Extractor}

\begin{figure}
    \centering
    \includegraphics[width=0.6\linewidth]{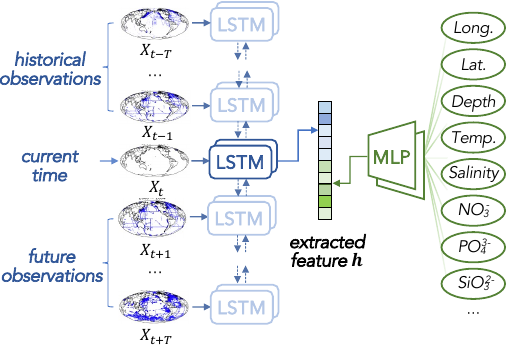}
    \caption{The detailed architecture of feature extractor.}
    \label{fig:feature-extractor}
\end{figure}

In our \textsc{OxyGenerator} feature extractor, we integrate historical ocean dissolved oxygen observations and multivariate oceanographic data. Figure \ref{fig:feature-extractor} illustrates the schematic diagram of our feature extractor, where we employ bidirectional LSTM to capture changes in historical and future ocean dissolved oxygen observations, achieving a "look back and look ahead" capability. For multiple physical and biogeochemical variables such as temperature, salinity, nitrate, phosphate, chlorophyll, etc., we employ a multi-layer perceptron to capture their nonlinear interconnections.
Ultimately, we concatenate the features from the two components to form a unified representation, which is utilized for the subsequent zoning-varying graph message-passing.

\subsection{Algorithm}
\label{appendix:algorithm}
In this section, we provide a detailed introduction to the algorithm of our proposed \textsc{OxyGenerator} in Algorithm \ref{alg:oxy}. 

\begin{algorithm}[tbh]
   \caption{Optimization algorithm of \textsc{OxyGenerator}}
   \label{alg:oxy}
\begin{algorithmic}[1]
   \REQUIRE Observed data $\textbf{X}$, geographical factor $\textbf{F}^{\text{Geo}}$, environmental factor $\textbf{F}^{\text{Env}}$
   \ENSURE \textsc{OxyGenerator} $\mathcal{M}$ with parameter $\theta^{*}, \phi^{*}$
   \STATE $\theta, \phi \leftarrow$ random initialization
   \STATE Split training data $\mathcal{B}_{\text{train}}$ into different batches via $N_{\text{area}}$ areas.
   \COMMENT{For each batch of data, it has a corresponding area ID.}
   \STATE Initialize the iteration numbers per area $T_{\text{area}} = [1/{N_{\text{area}}}, \cdots, 1/{N_{\text{area}}}]$
   \WHILE{not converged or max. epochs not reached}
        \FOR{a batch of training data ${B}_i$ from $\mathcal{B}_{\text{train}}$}
            \STATE Determine the area ID $n$ of ${B}_i$ and corresponding iteration number $T_n = T_{\text{area}}[n]$.
            \FOR{iteration number $T_n$}
                \STATE Calculate loss $\mathcal{L}^{(train)}_{\textsc{OxyGenerator}}$ via Equation \ref{eq:loss_oxy}.
                \STATE Update the model parameter $\theta, \phi$.
            \ENDFOR
        \ENDFOR
        \STATE Update the iteration numbers per area via average validation loss on different areas, i.e. $T_{\text{area}} = \text{softmax}\left[\mathcal{L}^{(val)}_{\textsc{OxyGenerator}}\right]$.
   \ENDWHILE
   \STATE Return the optimized model parameter $\theta^{*}, \phi^{*}$.
\end{algorithmic}
\end{algorithm}

\section{Experiment Details}
\label{appendix:exp-details}
\subsection{Evaluation Metrics}

In our experiments, we employed four metrics for performance evaluation, namely Root Mean Square Error (RMSE), Mean Absolute Percentage Error (MAPE), Mean Absolute Error (MAE), and Coefficient of Determination ($R^2$).

\begin{itemize}
    \item \textbf{Root Mean Square Error (RMSE)}: RMSE calculates the square root of the average of the squared differences between predicted and actual values, which is calculated as follows:
    \begin{equation*}
        \text{RMSE} = \sqrt{\frac{1}{n}\sum_{i=1}^{n} (X_i^{(obs)} - \hat{X}_i)^2}.
    \end{equation*}
    \item \textbf{Mean Absolute Percentage Error (MAPE)}: MAPE calculates the percentage difference between predicted and actual values, averaging these differences across all data points. The formula for calculating MAPE is as follows:
    \begin{equation*}
        \text{MAPE} = \frac{1}{n} \sum_{i=1}^{n} \left|\frac{X_i^{(obs)} - \hat{X}_i}{X_i^{(obs)}}\right| \times 100\%.
    \end{equation*}
    \item \textbf{Mean Absolute Error (MAE)}: MAE measures the average absolute differences between predicted and actual values. The formula for calculating MAE is as follows:
    \begin{equation*}
        \text{MAE} = \frac{1}{n} \sum_{i=1}^{n}|X_i^{(obs)} - \hat{X}_i|.
    \end{equation*}
    \item \textbf{Coefficient of Determination ($R^2$)}: $R^2$ is a statistical measure that represents the proportion of the variance in the dependent variable that is explained by the independent variables in a regression model. It is often used in the context of linear regression analysis.
    \begin{equation*}
        R^2 = 1 - \frac{\sum_{i=1}^{n} (X_i^{(obs)} - \hat{X}_i)^2}{\sum_{i=1}^{n} (X_i^{(obs)} - \bar{X}^{(obs)})^2}
    \end{equation*}
\end{itemize}

\subsection{Analysis on the Reconstruction of Deoxygenation in the Black Sea}
\label{appendix:black-sea}
When we examine the performance of various methods over time for reconstructing ocean deoxygenation, we observe a significant discrepancy among three numerical simulation methods during the period from 1923 to 1928. Furthermore, we identify that the major source of this substantial reconstruction error is the occurrence of highly inaccurate estimates in the Black Sea region, leading to a severe underestimation of the dissolved oxygen concentration in the Black Sea. Figure \ref{fig:black-sea} illustrates a scatter plot depicting the relationship between observed values and reconstructed values from different methods. It can be observed that many data points with low dissolved oxygen concentrations in the observational data are estimated as relatively high dissolved oxygen concentrations in the modeled data.

\begin{wrapfigure}{r}{0.35\textwidth}
  \vspace{-3mm}
  \centering
  \includegraphics[width=0.35\textwidth]{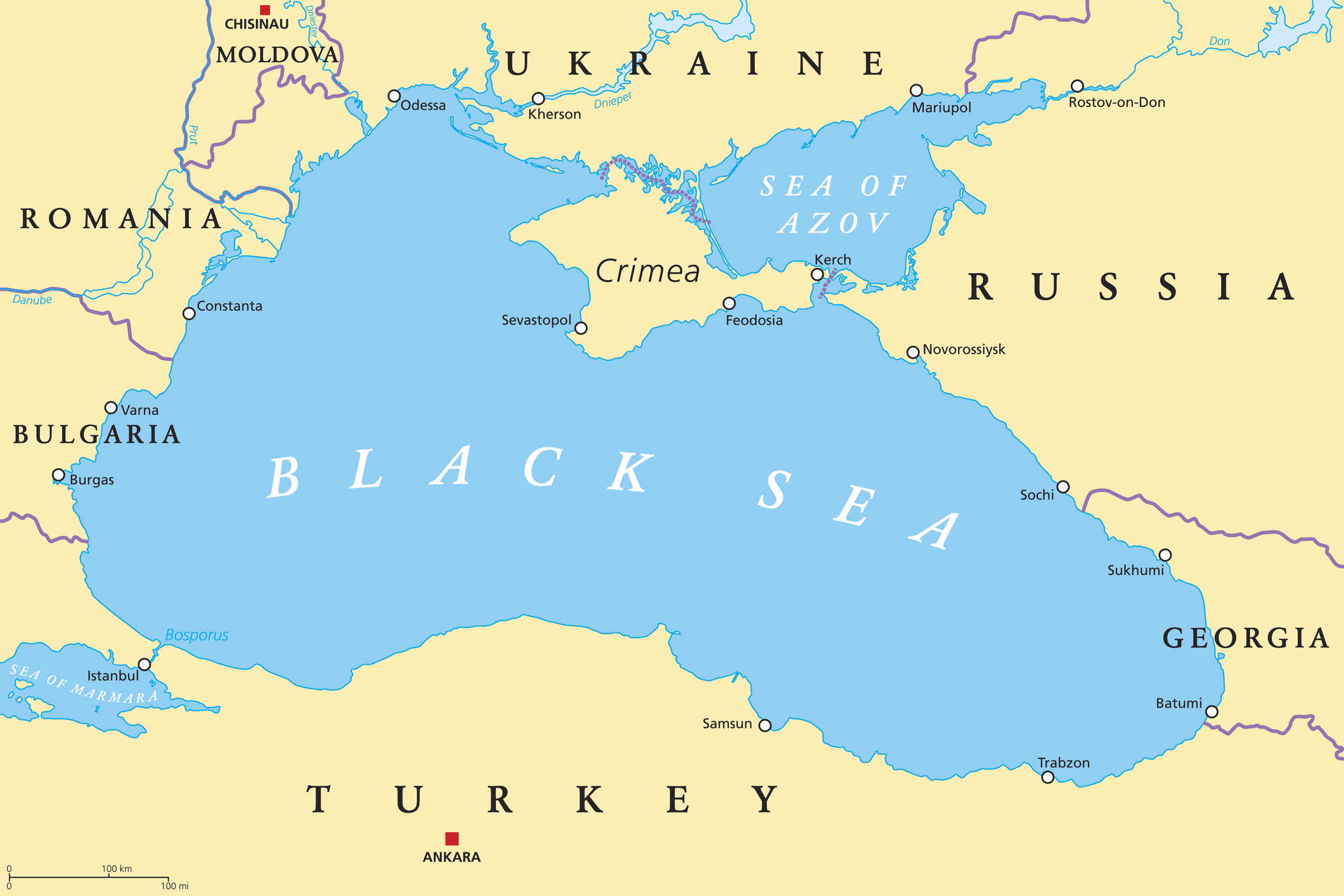}
  \caption{Map of the Black Sea.\protect\footnotemark[5]}
  \label{map:black-sea}
\end{wrapfigure}
\footnotetext[5]{Image source: iStock, Image ID: 802538470, Credit: PeterHermesFurian. Access from: \href{https://www.istockphoto.com/vector/black-sea-and-sea-of-azov-region-political-map-gm802538470-130198625}{https://www.istockphoto.com/vector/black-sea-and-sea-of-azov-region-political-map-gm802538470-130198625}}

Further discussions with oceanographers reveal that the Black Sea was once the largest natural dead zone (low-oxygen area). Oxygenated water is only found in the upper portion of the sea, where the Black Sea's waters mix with the Mediterranean Sea that flows through the shallow Bosporus strait. The north-west shelf of the Black Sea has suffered well-documented declines in biodiversity since the 1960s, and by the 1990s was considered a dead zone with virtually no sign of macroscopic epibenthic life. Therefore, numerical simulations based on the global climate system are unable to accurately depict the extreme deoxygenation in the Black Sea. Data-driven methods, on the other hand, prove effective in estimating extreme conditions by learning from historical observations and employing zoning-varying message passing.

\begin{figure}[h]
    \centering
    \includegraphics[width=\linewidth]{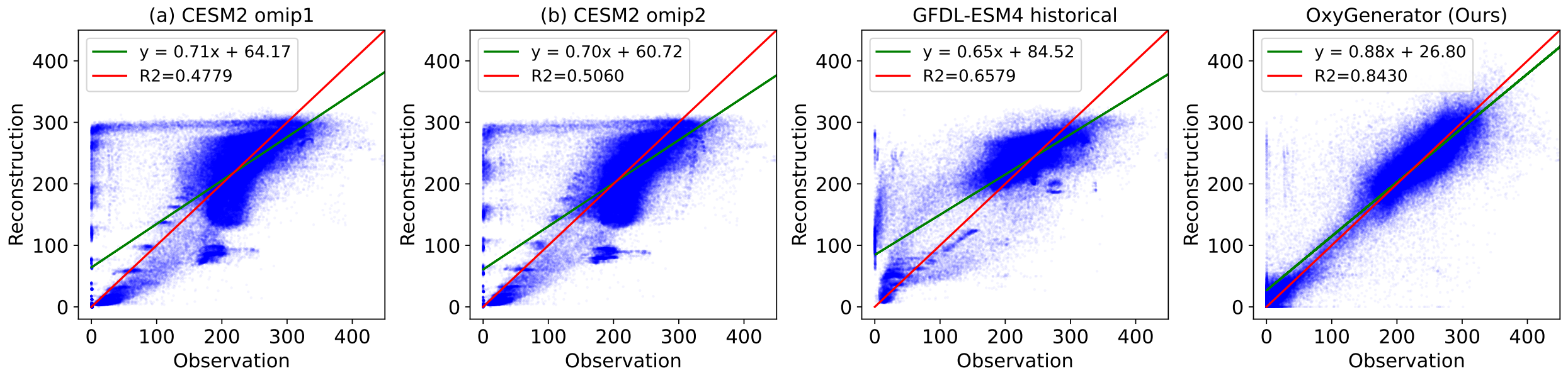}
    \caption{Performance Comparison of deoxygenation reconstruction of different numerical simulation methods and \textsc{OxyGenerator} in the Black Sea.}
    \label{fig:black-sea}
\end{figure}

\subsection{Comparison of Reconstruction Performance with Vertical Depth}
\label{appendix:vertical}

\begin{figure}
    \centering
    \includegraphics[width=0.9\linewidth]{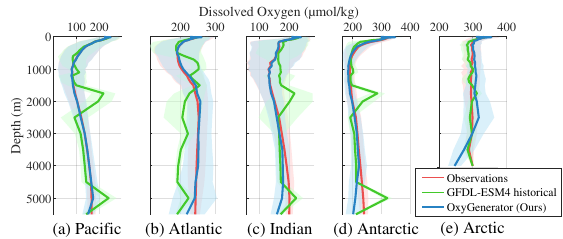}
    \caption{We divide the global ocean into five oceans and plot the changes of reconstruction results of our \textsc{OxyGenerator}, GFDL-ESM4 historical, and actual observations at different depths.}
    \label{fig:exp-depth}
\end{figure}

Figure \ref{fig:exp-depth} depicts the variation of reconstruction performance as the ocean depth changes. We visualize the observed values of dissolved oxygen, \textsc{OxyGenerator} reconstruction values, and GFDL-ESM4 historical numerical simulation results by partitioning them across the five oceans. The dissolved oxygen concentration in seawater generally exhibits a pattern of initially decreasing followed by an increase. We observe a high concordance between the reconstruction results of our method and the observed values, demonstrating consistency with the corresponding patterns. In contrast, the results based on the GFDL-ESM4 historical numerical simulation method exhibit significant fluctuations across different oceans, indicating a relatively high variability. This suggests that the modeling of simulation method in vertical depth layers lacks comprehensive consideration.

\subsection{Reconstruction Performance of Different Deep Learning Models}
\label{appendix:deep-learning-model}

To further compare and validate the effectiveness of our proposed method, we conduct comparisons with a series of deep learning models as follows:

\begin{itemize}
    \item \textbf{MLP (all)}: We use a multi-layer perceptron model for regression, taking all variables as input, including historical and future dissolved oxygen observation time sequences, geographical coordinates and environmental factors.
    \item \textbf{MLP (time series)}: Excluding environmental factors, we employ the time series of dissolved oxygen and geographical coordinates as inputs for the multi-layer perceptron.
    \item \textbf{MLP (environment)}: Excluding DO time series, we employ the environmental factors and geographical coordinates as inputs for the multi-layer perceptron.
    \item \textbf{GCN}: Graph Convolutional Network(GCN)~\cite{GCN} is a deep learning framework that leverages graph structure, utilizing graph-based convolutions to capture relations between nodes.
    \item \textbf{GAT}: Graph Attention Network (GAT)~\cite{DBLP:conf/iclr/VelickovicCCRLB18} is a type of neural network  designed for processing graph-structured data. GAT leverages attention mechanisms to dynamically weight the importance of neighboring nodes, allowing it to effectively capture complex relationships and dependencies within graph-structured data.
    \item \textbf{GraphSAGE}: GraphSAGE~\cite{DBLP:conf/nips/HamiltonYL17} is a graph neural network framework designed for scalable and inductive learning on large-scale graph-structured data, where it leverages neighborhood sampling and aggregation strategies to generate informative node embeddings.
    
    \item \textbf{MLP + GCN}: We use a combination of MLP and GCN for regression, using MLP as the feature extractor, taking all variables as input, including dissolved oxygen observation time sequences, geographical coordinates and environmental factors. Then the extracted features are passed through GCN for message passing and aggregation.

    \item \textbf{MLP + GAT}: The experimental setting is similar to LSTM + GCN, but replacing the graph neural network with Graph Attention Network.

    \item \textbf{MLP + GraphSAGE}: The experimental setting is similar to LSTM + GCN, but replacing the graph neural network with GraphSAGE.

    \item \textbf{BiLSTM}: Bidirectional Long Short-Term Memory Network (BiLSTM) is a type of recurrent neural network (RNN) architecture designed to effectively capture and remember long-range dependencies in sequential data for time series analysis. Therefore, we use Bidirectional Long Short-Term Memory Network as the feature extractor of dissolved oxygen time series, and the other factors are still feature extracted using multi-layer perceptron model, then concatenate the two outputs for regression.

    \item \textbf{Transformer}: Transformer is an attention-based method that efficiently capture contextual information from input sequences, leading to superior performance in tasks such as machine translation and language understanding.
    The experimental setting is similar to BiLSTM, except that the feature extractor of time series is changed to Transformer.
    
\end{itemize}

\begin{table}[]
\caption{Performance comparison with various machine learning methods.}
\label{tab:exp-ai-models}
\resizebox{\textwidth}{!}{%
\begin{tabular}{@{}lcccccccc@{}}
\toprule
\multirow{2}{*}{Benchmark}               & $k=1$       & $k=2$       & $k=3 $      & $k=4$       & \multicolumn{4}{c}{Average Performance}               \\ \cmidrule(l){2-9} 
                                         & MAPE & MAPE & MAPE & MAPE & MAPE  & R2             & RMSE       & MAE        \\ \midrule
MLP       (all)                            & 22.14     & 24.87     & 22.98     & 22.12     & 23.03\scriptsize{±1.29} & 0.8641\scriptsize{±0.0135}  & 30.89\scriptsize{±0.95} & 22.08\scriptsize{±1.01} \\
 MLP (time series)        & 31.28     & 29.69     & 34.43     & 24.81     & 30.05\scriptsize{±4.01} & 0.8127\scriptsize{±0.0065}  & 36.52\scriptsize{±1.17} & 26.90\scriptsize{±0.73} \\
MLP (environment) & 24.40     & 23.19     & 25.90     & 20.85     & 23.59\scriptsize{±2.13} & 0.8524\scriptsize{±0.0127}  & 32.39\scriptsize{±0.99} & 23.69\scriptsize{±1.04} \\ \midrule
GCN                                      &  47.04    &   51.24  &  42.73    &  29.08    & 42.70\scriptsize{±9.27} & 0.6085\scriptsize{±0.0282}  & 49.30\scriptsize{±2.24} & 37.04\scriptsize{±2.13}\\
GAT                                      & 40.69     & 44.57     & 43.94     & 29.80     & 39.75\scriptsize{±6.84} & 0.6930\scriptsize{±0.0333}  & 46.68\scriptsize{±1.45} & 35.32\scriptsize{±1.54 }\\
GraphSAGE  & 42.96     & 45.49     & 48.91     & 32.76     & 42.53\scriptsize{±6.95} & 0.6750\scriptsize{±0.0304}  & 48.09\scriptsize{±2.61 } & 36.33\scriptsize{±2.17 }\\
MLP + GCN & 31.00  & 32.74 & 27.62 &31.73 &30.63\scriptsize{±2.15} & 0.7772\scriptsize{±0.0301} &39.76\scriptsize{±2.72} & 29.31\scriptsize{±2.84}\\
MLP + GAT & 28.39  & 27.64 & 26.49 &27.80 & 27.58\scriptsize{±0.79} &0.8029\scriptsize{±0.0054}  &37.47\scriptsize{±0.90} &27.74\scriptsize{±0.69} \\
MLP + GraphSAGE & 25.74 & 28.43 & 26.39 & 20.51 &  25.27\scriptsize{±3.37} &  0.8382\scriptsize{±0.0082}  & 33.95\scriptsize{±1.54}  & 24.92\scriptsize{±1.47}\\\midrule
BiLSTM & 22.88 & 21.27 & 23.58 & 21.19&22.37\scriptsize{±1.19} & 0.8724\scriptsize{±0.0017} & 30.14\scriptsize{±0.58} & 21.58\scriptsize{±0.53}\\
Transformer & 24.07 & 25.83 & 27.38 & 19.09 & 24.25\scriptsize{±3.61} &0.8607\scriptsize{±0.0091} &31.49\scriptsize{±1.66} & 22.59\scriptsize{±1.20} \\


\midrule

\textsc{OxyGenerator} (Ours) & \textbf{14.72} & \textbf{13.48} & \textbf{15.72} & \textbf{13.20} & \textbf{14.28\scriptsize{±1.16}} & \textbf{0.9026\scriptsize{±0.0072}} & \textbf{26.31\scriptsize{±1.23}} & \textbf{17.57\scriptsize{±1.10}} \\

\bottomrule
\end{tabular}%
}
\end{table}

Table \ref{tab:exp-ai-models} records the performance of different methods in a 4-fold cross-testing and the average performance across all folds. The results show that \textsc{OxyGenerator} outperforms all baseline deep learning models, with a performance improvement of about 36.16\% in MAPE metric. Meanwhile, the experiments show that MLP can achieve similar performance to CMIP6, and the addition of time series and environmental variables will further improve the performance. Graph neural network such as GAT and GraphSAGE performs poorly, which due to the incapability of classical GNN models to deal with the spatio-temporal heterogeneity.
In addition, in terms of selecting the feature extractor for the time series, the best performance was achieved using BiLSTM, which outperformed MLP and Transformer in terms of feature capture for the time series. The reason for the poor performance of Transformer is that  Attention mechanism struggles to deal a large number of missing values in time series.



\subsection{Comparison of \textsc{OxyGenerator}'s Adaptive Zoning with World Ocean Database 2018.}
\label{appendix:zoning}

\begin{figure}[htbp]
    \begin{minipage}{0.5\textwidth}
        \centering
        \includegraphics[width=\linewidth]{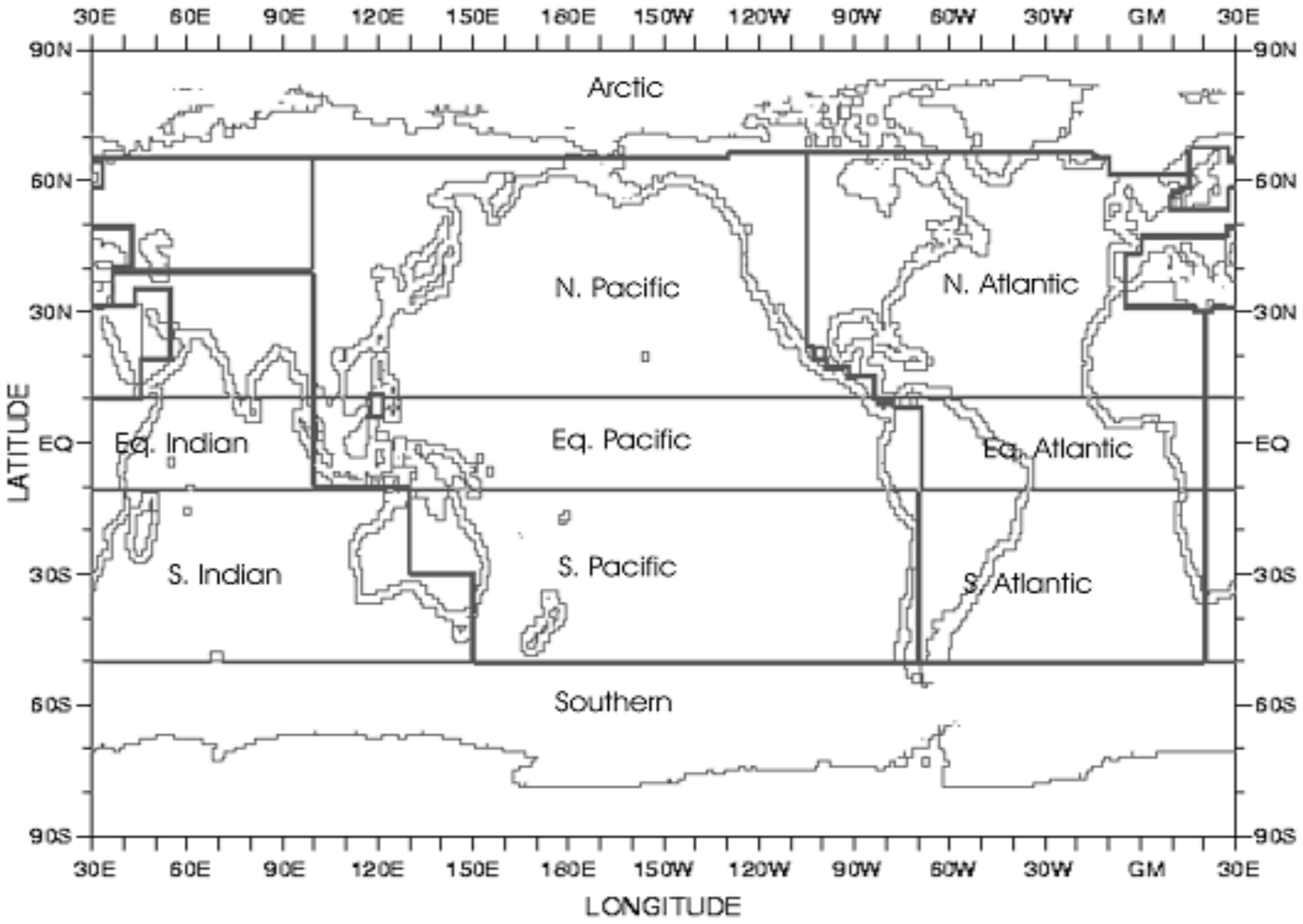}
        \caption{Geographic boundaries of ocean basin definitions in WOD18.}
            \label{fig:WOD-zoning}
    \end{minipage}%
    \hfill
    \begin{minipage}{0.5\textwidth}
        \centering
        \captionsetup{type=table}  
        \caption{Detailed list of zoning in WOD18}
        \label{tab:wod-tab}
        \begin{tabular}{clcl}
        \toprule
        \textbf{id} & \textbf{area} & \textbf{id} & \textbf{area} \\
        \midrule
        1. & Climatology Range     & 16. & Equatorial Indian    \\
        2. & North Atlantic        & 17. & Coastal Eq Indian    \\
        3. & Coastal N Atlantic    & 18. & South Indian         \\
        4. & Equatorial Atlant     & 19. & Coastal S Indian     \\
        5. & Coastal Eq Atlant     & 20. & Antarctic            \\
        6. & South Atlantic        & 21. & Arctic               \\
        7. & Coastal S Atlantic    & 22. & Mediterranean        \\
        8. & North Pacific         & 23. & Black Sea            \\
        9. & Coastal N Pac         & 24. & Baltic Sea           \\
       10. & Equatorial Pac        & 25. & Persian Gulf         \\
       11. & Coastal Eq Pac        & 26. & Red Sea              \\
       12. & South Pacific         & 27. & Sulu Sea             \\
       13. & Coastal S Pac          &        &                  \\
       14. & North Indian            &     &                     \\
       15. & Coastal N Indian        &     &                    \\
       \bottomrule
    \end{tabular}
    \captionsetup{labelformat=default} 
    \vfill 
    \end{minipage}
\end{figure}


\textbf{Introduction to the WOD18 zoning strategy.}  In oceanographic studies, many studies divide the global ocean into different zones~\cite{fay2014global,Reygondeau2020ClimateCE,Sonnewald2020ElucidatingEC} for deeper study of complex ocean processes. Among them, the World Ocean Database 2018 (WOD18)~\cite{wod2018} zoning is the most popular and broadly used. The WOD18 divides the world's oceans into 27 different zones based on hand-crafted rules, as shown in Figure \ref{fig:WOD-zoning} and Table \ref{tab:wod-tab}. 
The zoning strategy for WOD mainly relies on ocean basins.
Ocean basins are the basic geographic units of the global oceans, which indicate large bodies of water consisting of oceanic ridges, continental slopes and ocean basins. Ocean basin zoning aims to divide the global ocean into major regions according to different geographic, geological, and oceanographic features.

\textbf{Effect of Adpative Zoning via Spatio-Temporal HyperNetwork.} We compare the zoning result of hypernetwork in Figure \ref{fig:hypernetwork} with WOD18. We find that the result both maintains consistency with WOD18 and reflects a finer-grained ocean zoning.
\begin{itemize}
    \item \textbf{The adaptive zoning result shows consistent with WOD18 in many regions.} \textsc{OxyGenerator}'s zoning of the oceans is closely related to  latitudinal distribution, which is similar to the WOD18 and consistent with the spatial and temporal distribution characteristic of dissolved oxygen. Meanwhile, even without the hand-craft rules, \textsc{OxyGenerator} achieves a clear partition among different oceans. Especially below 450 meters depth, our zoning result is highly consistent with WOD18. More specifically, our method divides the Indian Ocean into two zones colored in light blue and blue, similarly corresponding to Equatorial Indian and South Indian in WOD18. The Atlantic Ocean is divided into three zones colored in light green, pink and orange, corresponding to North Altantic, Equratorial Atlantic and South Altantic in the WOD18.
    \item \textbf{The adaptive zoning result shows finer-grained ocean zoning.} Our method depicts a finer grained zoning that varies at different depths. 
    Rather than depth-independent zoning strategy in WOD, \textsc{OxyGenerator} is capable of performing adaptive zoning over different depth levels. Specifically, surface seawater above 100m are more affected by human activities and the DO distribution pattern becomes more complex. Therefore, \textsc{OxyGenerator} presents a more intricate zoning pattern in Figure \ref{fig:hypernetwork}(a-b), which is especially obvious in the Pacific and Atlantic Oceans. The deep seawater is more affected by ocean circulation in different geographic locations, hereby our \textsc{OxyGenerator} shows the adaptive zoning associated with oceanic boundaries. 
\end{itemize}

\section{Ocean Deoxygenation Reconstruction via \textsc{OxyGenerator}}

\begin{figure}[h]
    \centering
    \includegraphics[width=\linewidth]{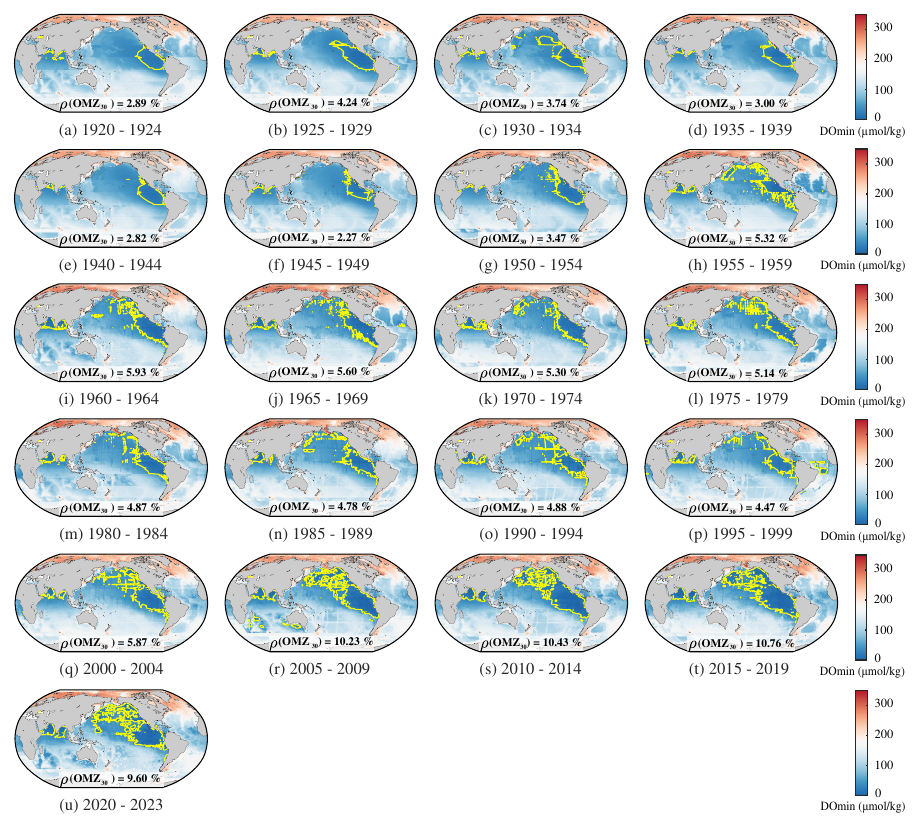}
    \caption{Global Minimum DO reconstructed  via \textsc{OxyGenerator} from 1920 to 2023.}
    \label{fig:global_omz30}
\end{figure}

In this section, we show the reconstruction results carried out by our proposed \textsc{OxyGenerator}. Figure \ref{fig:global_omz30} illustrates the minimum oxygen concentration every five years from 1920 to 2023. The yellow contour lines depict the oxygen minimization zones (OMZs) where the dissolved oxygen concentration is less than 30 $\mu$mol/kg, which we can further denoted as $\text{OMZ}_{\text{30}}$. We clearly observe a continuous increase in the proportion of $\text{OMZ}_{\text{30}}$ with the passage of years, indicating a growing impact of oceanic deoxygenation over time. 


\end{document}